\documentclass[letterpaper,twocolumn,10pt]{article}

\usepackage{usenix}
\usepackage{tikz}
\usepackage{amsmath,amssymb}
\usepackage{xspace}
\usepackage{booktabs}
\usepackage{algorithm}
\usepackage{algorithmic}
\usepackage{graphicx}
\usepackage{array}
\usepackage[table]{xcolor}
\usepackage{multirow}
\usepackage{colortbl}

\newcommand{\method}{\textsc{ARBITRA}\xspace}
\newcommand{\argora}{\textsc{ARGORA}\xspace}
\newcommand{\attack}{ATT\&CK\xspace}
\newcommand{\ttp}{TTP\xspace}
\newcommand{\cti}{CTI\xspace}

\definecolor{fpcolor}{RGB}{180,55,55}
\definecolor{tpcolor}{RGB}{55,90,170}
\newcommand{\fpobj}{%
  \textcolor{fpcolor}{\textbf{\ensuremath{\oplus}\,FP}}%
}
\newcommand{\tpobj}{%
  \textcolor{tpcolor}{\textbf{\ensuremath{\ominus}\,TP}}%
}

\definecolor{ctagcolor}{RGB}{0,90,160}  
\newcommand{\ctag}[1]{\textcolor{ctagcolor}{\textbf{(#1)}}}

\newcommand{\Acal}{\mathcal{A}}
\newcommand{\Bcal}{\mathcal{B}}
\newcommand{\Dcal}{\mathcal{D}}

\newcommand{\Xcal}{\mathcal{X}}
\newcommand{\Ycal}{\mathcal{Y}}
\newcommand{\Parse}{\mathsf{Parse}}

\begin{document}

\date{}

\title{\Large \bf \method: Robust Document-Level ATT\&CK Extraction via\\Multi-Expert Argumentation}


\author{
    Youngjin Jin \qquad Seungwon Shin\\
    School of Electrical Engineering, KAIST\\
    \texttt{\{ijinjin,claude\}@kaist.ac.kr}
}

\maketitle

\begin{abstract}
Cyber threat intelligence (\cti) reports describe adversary behavior in natural language, but converting them into structured MITRE \attack mappings remains difficult: existing methods decompose reports to sentence-level granularity for extraction, while modern LLMs can work with full reports in a single pass but with noticeable performance degradation. We present \method, which orchestrates multiple role-specialized LLM experts that propose candidate \attack entities from the full report, which are then verified through a per-candidate quantitative bipolar argumentation framework (QBAF) grounded in document evidence and rationales. We additionally construct reasoning-specialized corpora for finetuning so experts can act as effective participants during this verification process. Across \attack extraction benchmarks, \method reaches a stronger precision-oriented operating point than sentence-level pipelines, recent LLM-based extraction systems, and frontier models including GPT-5.5. Under adversarial perturbations simulating conflicting \ttp statements, cross-report contamination, and semantic poisoning, \method limits mean precision degradation to roughly half that of frontier-model baselines for a more robust extraction.
\end{abstract}

\section{Introduction}
\label{sec:introduction}

Cyber threat intelligence (\cti) reports are a primary medium through which security analysts communicate adversary behavior: what an adversary did, how it operated, and what defenders should anticipate. Unlike raw indicators of compromise, \cti reports typically describe higher-level tactics, techniques, and procedures (\ttp{}s) like persistence mechanisms, credential access, or malware capabilities. The MITRE \attack framework~\cite{mitre_attack} provides a widely adopted ontology for representing these behaviors and underpins downstream workflows such as threat hunting, detection engineering, incident triage, and threat attribution.

This practical utility of \cti, however, depends on whether analysts can convert long, unstructured reports into reliable structured knowledge. A single report may interleave concrete behavioral evidence, background description of an actor, analyst hypotheses, mitigation guidance, or even unrelated or contradictory statements. Mapping such text to \attack techniques is therefore not a keyword-recognition problem: a system must determine whether a technique is genuinely supported by report-level evidence.

Prior work has made substantial progress through rule-based systems, NER pipelines, graph construction, and learning-based approaches such as TTPDrill, AttacKG, and TTPXHunter~\cite{ttpdrill,attacKG,ttpxhunter}. However, a recent USENIX Security SoK~\cite{sok_ttp_extraction_2025} re-evaluated under a unified protocol on public resources such as TRAM2~\cite{tram} and AnnoCTR~\cite{annoctr} and found a performance ceiling no existing approach overcomes, with traditional NLP pipelines outperforming generative LLM approaches in realistic open-set settings. The SoK attributes this to ontology ambiguity: many \attack techniques differ only in intent and cannot be distinguished from a single sentence, yet the systems that perform well remain sentence-centric. Reading the full document is necessary but not sufficient: full-document LLM prompting and even frontier LLMs over-accept plausible but unsupported techniques and degrade further on misleading or contaminated content~\cite{sok_ttp_extraction_2025, mezzi2025unreliable}. Since false positives misdirect analyst attention and pollute detection systems, a practical extractor must read at document level, recover performance lost from single-pass prompting, and accept a candidate only when it is grounded in report evidence and contestable by an analyst.

To this end, we introduce \textbf{ARBITRA} (\textbf{A}rgumentative \textbf{R}easoning with \textbf{B}ipolar \textbf{I}nference for \textbf{T}hreat \textbf{R}eport \textbf{A}nalysis), a document-level extraction system built around argumentation-based multi-expert verification. \method is inspired by \argora~\cite{jin2026argora}, which provides a general mechanism for structured multi-expert argumentation, but is not well-suited for document-level, multi-label extraction. \method reformulates the problem as a two-phase candidate-verification procedure: multiple role-specialized LLM experts first read the full report and independently propose candidate MITRE entities, favoring coverage through expert diversity; each non-unanimous candidate is then recast as a claim for which experts generate supporting and attacking arguments based on their expertise, and the resulting quantitative bipolar argumentation framework (QBAF) decides acceptance. \method is thus a candidate-level verification layer atop document-level expert extraction.

A challenge with this design is that generic LLM experts may not be effective participants in these argumentation-based verification settings, as experts trained only to emit final labels produce justifications that are shallow or poorly aligned. We therefore construct reasoning-specialized supervised fine-tuning corpora from the training splits of the TRAM2 and AnnoCTR data and compare them against answer-only fine-tuning to test whether rationale-rich experts are beneficial for downstream candidate verification. Finally, we evaluate robustness under adversarial \cti conditions designed to test whether an extractor preserves precision when a report contains plausible but unsupported or conflicting evidence. 

In summary, our contributions are as follows:
\begin{itemize}
    \item We formulate document-level \attack extraction as a precision-oriented candidate-verification problem, motivated by the limitations of sentence-centric pipelines and naive full-document LLM extraction.
    \item We introduce \method, a two-phase document-level extraction framework that pairs full-report expert candidate generation with QBAF-based verification.
    \item We construct reasoning-based fine-tuned experts to leverage the benefits of the structured multi-expert verification design of \method.
    \item We evaluate \method against frontier LLMs (e.g., GPT-5.5), the SoK's best-performing sentence-level pipeline, and three recent extraction systems (TTPXHunter, CTINexus, TTPrint). \method is the only system to surpass the sentence-level pipeline on all three metrics on TRAM2 and attains the highest precision on both benchmarks.
    \item We introduce an adversarial \cti robustness evaluation suite and show that \method retains extraction precision under these conditions more reliably than frontier-model baselines.
\end{itemize}
\section{Related Work}
\label{sec:related_work}

\noindent
\textbf{Automated \ttp Extraction from \cti Reports.}
Early automated \ttp extraction systems recovered security behavior from unstructured \cti text using linguistic patterns, domain knowledge, and graph construction: TTPDrill combines NLP with security-specific inference \cite{ttpdrill}, AttacKG builds technique knowledge graphs from reports \cite{attacKG}, and TTPXHunter aggregates sentence--TTP predictions to the report level \cite{ttpxhunter}. The recent USENIX Security SoK by Büchel et al.~\cite{sok_ttp_extraction_2025} re-implements the core methods of these systems under a unified protocol and reaches conclusions that shape the motivation for our work: existing approaches hit a performance ceiling; classical NLP pipelines outperform embedding-based and generative LLM methods once the technique set is realistic and open; and the limiting factor is ontology ambiguity that cannot be resolved from sentence-level granularity. We adopt the SoK's findings and target the document-level, open-set regime in which it finds generative approaches weakest.

\noindent
\textbf{LLM-Based \cti Extraction.}
LLMs are an appealing alternative to sentence-level pipelines because they read longer contexts and emit structured outputs directly. CTIBench~\cite{ctibench} and CTIConnect~\cite{ctiarena} evaluate existing LLMs on \cti tasks and confirm that frontier models lead most \attack-oriented extraction. System-level efforts such as CTINexus~\cite{ctinexus} and ThreatPilot~\cite{intelex} use in-context learning and LLM-driven pipelines to turn unstructured reports into structured intelligence. Long-context extraction however also introduces new failure modes: LLMs can be distracted by irrelevant context \cite{shi_irrelevant_context}, and recent evaluations find LLMs inconsistent and overconfident on real-size reports which lead to precision degradation \cite{mezzi2025unreliable}. The SoK's document-level experiments suggest a large recall degradation compared to sentence-level granularity~\cite{sok_ttp_extraction_2025}. \method addresses both sides of this trade-off. A union over role-specialized experts recovers the recall that a single pass loses, and per-candidate argumentation filters the union so that precision is also retained.

A recent, concurrent work called TTPrint~\cite{ttprint} decomposes a report into atomic behaviors, anchors each to a lexically localized evidence span, and accepts candidate techniques whose single-verifier confidence exceeds a threshold. \method shares the premise that candidate generation and acceptance should be decoupled, but differs in how acceptance is decided: TTPrint scores each candidate using a single LLM verifier, whereas \method adjudicates each disputed candidate through structured argumentation from multiple role-specialized experts over the full document. Our ablation in Section~\ref{subsec:eval_verifier} shows that collapsing this step to a single-call verifier degrades F1 by up to 15 pp.

\noindent
\textbf{Reasoning Fine-Tuning.}
Reasoning-elicitation based works such as chain-of-thought (CoT)~\cite{NEURIPS2022_9d560961}, instruction tuning~\cite{ouyang2022training_instructgpt,wang2023self_instruct}, and rationale-based training~\cite{zelikman2022star,hsieh2023distilling_step_by_step} show that generation or supervision of intermediate steps rather than outcomes alone can matter for reliability \cite{lightman2023letsverify}. These motivate the hypothesis that even small models can become stronger task specialists when trained to expose the reasoning process, not only the final answer. We raise a narrower question: \textit{does reasoning supervision help when the same model also participates in argumentation}? \method studies this by constructing both answer-only and reasoning-specialized fine-tuning datasets from TRAM2 and AnnoCTR; to our knowledge this comparison has not been isolated inside a structured multi-expert verification pipeline.

\noindent
\textbf{Multi-LLMs and Multi-Agents.}
A growing body of work improves LLM reliability by sampling multiple reasoning paths or coordinating multiple agents: self-consistency aggregates several CoT paths by voting \cite{wang2022self}; multi-agent debate has models critique one another \cite{10.5555/3692070.3692537,liang-etal-2024-encouraging}; and multi-expert prompting and mixture-of-agents use multiple specialized or layered calls \cite{long-etal-2024-multi-expert,wang2024mixture}. Most recently, ARGORA organizes multi-experts into explicit argumentation graphs for decision-level reasoning and diagnostics \cite{jin2026argora}. These methods motivate expert diversity but mostly aggregate at the answer level or rely on unstructured transcripts. ARGORA is a closer precursor but its setting assumes a small set of competing alternatives rather than the verification setting necessary for document-level, multi-label \attack extraction. As such, \method allows experts to disagree on each technique without invalidating the rest of their output.

\noindent
\textbf{Adversarial \cti.}
Security text is not guaranteed to be clean: false-flag artifacts complicate attribution \cite{skopik_false_flag}, LLMs can generate fake \cti-like text that poisons downstream defenses \cite{ranade_fake_cti}, models are easily distracted by irrelevant context \cite{shi_irrelevant_context}, and cybersecurity NER systems are vulnerable to textual adversarial attacks \cite{jiang_adv_cyber_ner}. Our adversarial evaluation is designed around these observations and assesses robustness on realistic forms of \cti noise and manipulation, and is the first work to define an adversarial CTI threat model for MITRE extraction and instantiate full-report perturbations.
\section{Problem Definition}
\label{sec:def-problem}

\begin{figure*}[!htbp]
  \centering
  \includegraphics[
    width=0.95\linewidth,
    keepaspectratio,
    trim=3mm 6mm 3mm 5mm,
    clip
  ]{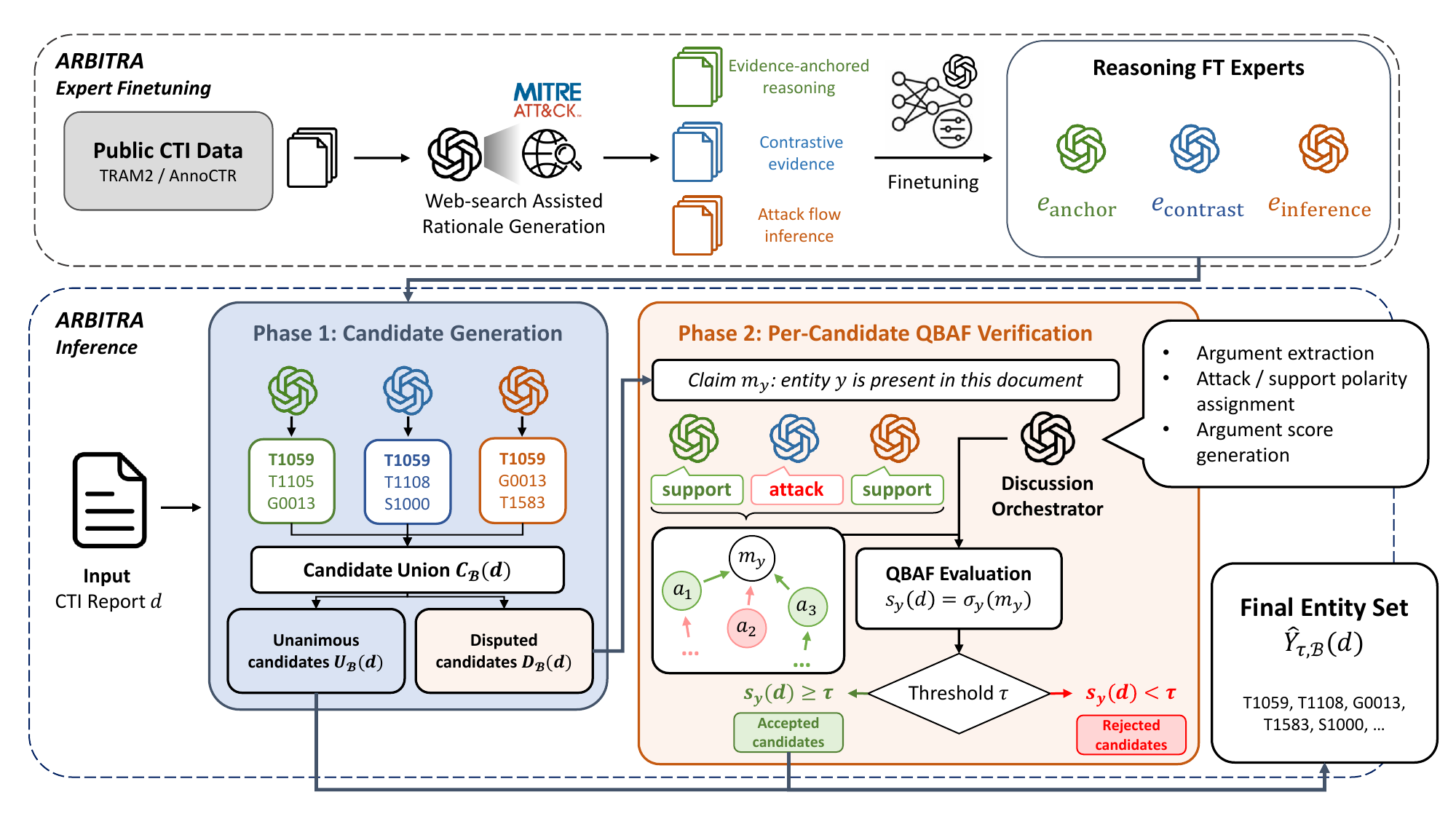}
  \caption{Overview of \method as described in Section~\ref{sec:methodology}.}
  \label{fig:arbitra-overview}
\end{figure*}

\subsection{Background and Motivation}
\label{subsec:background}

\noindent
\textbf{Precision and Contestability.}
False positives carry a heavier cost beyond missed recall in CTI: an unsupported technique can send analysts after irrelevant behaviors or pollute a knowledge base that later systems treat as trusted input. This motivates a precision-oriented view of document-level extraction. One such method would be to allow uncertain candidates to undergo a focused verification step. This verification process should be \emph{contestable}; simple judge-based verification mechanisms fall short of these requirements. A single LLM confidence score is uncalibrated and opaque: it collapses all considerations into a scalar without separating evidence in favor of a mapping from evidence against it, and it provides no handle for an analyst to inspect which evidence drove the decision. We describe this as one of the key challenges to solve in Section~\ref{subsec:def-challenges}, and design our methodology around this challenge as shown in Section~\ref{sec:methodology}.

\noindent
\textbf{Noisy and Adversarial CTI Inputs.}
\cti reports can carry false or misleading technique attributions (as mentioned in Section~\ref{sec:related_work}), can be contaminated by passages aggregated from unrelated incidents, and can describe genuine behavior in paraphrased language that suppresses entity recognizability. We formalize these degradation classes as a threat model in Section~\ref{subsec:def-threatmodel} and construct adversarial variants in Section~\ref{sec:evaluation_adversarial}.

\subsection{Task: Document-Level CTI Extraction}
\label{subsec:def-task}

We study \textit{document-level} \attack entity extraction from \cti reports. Given a full \cti report $d$, the task is to produce a set of normalized MITRE \attack entity labels $\widehat{Y}_{\mathcal{B}}(d)$ that approximates the gold label set $Y_{\mathcal{B}}(d)$ under a benchmark-specific label space $\mathcal{Y}_{\mathcal{B}}$. We use \emph{MITRE \attack entity} to refer collectively to the benchmark-specific label types considered in this paper: \attack techniques and sub-techniques, tactics, groups, and software.

\subsection{Threat Model: Adversarial CTI Reports}
\label{subsec:def-threatmodel}

Our threat model is grounded in documented \cti cases of adversarial manipulation and ordinary noise. The 2018 Olympic Destroyer operation is a canonical example: the Sandworm-attributed malware was deliberately crafted to resemble Lazarus Group samples as a false flag intended to misdirect attribution~\cite{kaspersky_olympic_falseflag,doj_sandworm}. More recently, adversarial rewriting can suppress \attack-recognizable cues while preserving the underlying behavior~\cite{jiang_adv_cyber_ner}. Similar degradation also arises non-adversarially through feed aggregation, analyst errors, and reporting-style variation. We taxonomize three such perturbation types below.

\noindent\textbf{Target system.}
The target system is an automated \attack extraction pipeline that ingests \cti reports and outputs structured \attack entity sets for downstream use such as intrusion detection. The pipeline is exposed to the content of each ingested report and trusts that report as its primary input.

\noindent\textbf{Adversary goal.}
The adversary aims to degrade the quality of the extraction output by manipulating CTI reports. We consider two complementary degradation objectives: \emph{false-positive injection} (\fpobj{}), in which the
adversary causes the extractor to accept \attack entities that are not supported by the underlying evidence; and \emph{true-positive suppression} (\tpobj{}), in which the adversary causes the extractor to miss
entities supported by the original document.

\noindent\textbf{Adversary capabilities.}
The adversary can modify \cti report content before it enters the extraction pipeline. The original incident and its gold label set are left unchanged; only the document text is modified.
\begin{itemize}
  \item \textit{Conflicting TTP statements}~(\fpobj).
  The adversary inserts plausible text that attributes
  adversary behavior to unsupported \attack entities. This models false-flag artifacts, deliberate
  misattribution, or the injection of fabricated analysis into shared threat intelligence~\cite{skopik_false_flag,ranade_fake_cti}.

  \item \textit{Cross-report contamination}~(\fpobj).
  The adversary interleaves passages from unrelated \cti documents into the target report. This models noisy aggregation from heterogeneous threat feeds or manipulation of shared intelligence repositories~\cite{johnson2016guide_nist_cti}.

  \item \textit{Semantic poisoning}~(\tpobj).
  The adversary paraphrases or rewrites entity-signaling language in labeled reports. This models adversarial obfuscation of \attack-recognizable phrasing~\cite{jiang_adv_cyber_ner} or distribution shift between reporting styles.
\end{itemize}

\noindent\textbf{Adversary knowledge.}
The adversary is assumed to have \emph{black-box} access to the extraction system: it knows the system's capabilities (e.g., \attack-based extraction) but does not know the system's model weights, prompts, fine-tuning data, or internal architecture. The adversary operates purely at the text level and requires no knowledge of model internals. Our threat model is scoped to realistic text-level manipulations of \cti reports that an adversary could deploy through compromised intelligence sources.

\noindent\textbf{Non-adversarial occurrences.}
The non-adversarial instantiations of all three classes are grounded in documented CTI practice. For \textit{conflicting TTP statements}, manual TTP mapping is characterized as being error prone even under careful conditions~\cite{baker_tram_2024}, and CISA's \attack mapping guidance explicitly recognizes that insufficient behavioral evidence leads to misattribution~\cite{cisa_mitre_attack_mapping}. For \textit{cross-report contamination}, the SolarWinds/SUNBURST campaign was labeled under at least four distinct vendor-assigned identifiers (UNC2452, NOBELIUM, Dark Halo, and SolarStorm) before attribution was consolidated~\cite{attack_solarwinds_c0024}; an empirical study of multi-report \attack extraction across 90 reports demonstrates that multi-report aggregation changes extraction behavior and up to 33.3\% of misclassifications involve semantically similar techniques~\cite{beyond_single_reports}. For \textit{semantic poisoning}, reporting style varies substantially across vendors, with some using \attack-recognizable terminology and others describing the same behavior in narrative or proprietary language, producing surface-level signal suppression indistinguishable from deliberate obfuscation.

\subsection{Challenges}
\label{subsec:def-challenges}

We present three challenges that prior approaches do not jointly address as follows.

\noindent\textbf{\ctag{C1}: Precision control under full-document extraction.} When a model reads the full report and generates entity labels in a single pass, it has no mechanism to distinguish candidate discovery from candidate acceptance. Hallucinated labels are easily generated because suggestive attack patterns or terminology can all trigger an entity prediction.

\noindent\textbf{\ctag{C2}: Expert rationales for verification-based design.} The two-phase design introduced to address \ctag{C1} requires experts to produce grounded support and attack arguments, not just final labels. LLMs trained only on label prediction produce rationales that are shallow or poorly aligned with the evidential requirements of \attack mapping and are unsuited for a contestable verification step.

\noindent\textbf{\ctag{C3}: Robustness to adversarial report content.} Under the threat model of Section~\ref{subsec:def-threatmodel}, a report may contain plausible but unsupported technique claims, irrelevant \attack-relevant passages, or weakened surface cues for genuinely supported techniques. An extractor that makes document-level decisions without a verification step is vulnerable to all three classes.

\subsection{Design Goal}
\label{subsec:def-goal}

\method is designed to address \ctag{C1}--\ctag{C3} within a single unified framework. To address \ctag{C1}, \method decouples candidate generation from candidate acceptance through a two-phase pipeline: Phase~1 builds a broad union of expert-proposed candidates, and Phase~2 subjects each non-unanimous candidate to evidence-grounded QBAF verification. To address \ctag{C2}, its experts are fine-tuned on reasoning-specialized corpora that contain grounded rationales. To address \ctag{C3}, the QBAF verification step provides a mechanism for rejecting candidates that cannot accumulate sufficient argumentative support from the report evidence, making the system inherently harder to mislead with adversarial content. In the rest of the paper, we mark with \ctag{C1}--\ctag{C3} the design decisions and experimental results that address each
challenge to trace how each challenge is resolved.
\section{\method}
\label{sec:methodology}

This section describes \method. An illustrated overview of \method is shown in Figure~\ref{fig:arbitra-overview} for reference.

\noindent
\textbf{Preliminaries.}
We use quantitative bipolar argumentation frameworks (QBAFs) \cite{BaroniRagoToni2019Spectrum,Rago2016DFQuAD,Kampik2024Contribution} to represent each per-candidate verification discussion. QBAFs provide semantics for computing acceptance strength from structured support and attack arguments necessary for transparent, contestable candidate verification. Previous multi-expert argumentation systems such as \argora \cite{jin2026argora} apply QBAFs to single-choice decision problems, where a small set of competing expert arguments is compared and one answer is selected. \method instead instantiates per-candidate QBAF verification within a multi-label document-level extraction pipeline, which requires a fundamentally different formulation in which each non-unanimous candidate is adjudicated independently. Formally, a QBAF is a tuple
\begin{equation}
\label{eq:method_qbaf}
  Q = \langle \Acal, R^{-}, R^{+}, w \rangle,
\end{equation}
where $\Acal$ is a finite set of textual arguments, $R^{-}\subseteq \Acal\times\Acal$ and $R^{+}\subseteq \Acal\times\Acal$ are directed attack and support relations, and $w:\Acal\rightarrow[0,1]$ assigns each argument a base score. We represent each QBAF as a rooted directed tree whose edges are oriented from child to parent. For a node $a$, let $\operatorname{ch}(a)=\{c_1,\ldots,c_n\}$ denote its children and let $\pi(a)\in\{-1,+1\}^n$ denote the corresponding attack/support polarity vector.

Quantitative semantics computes a final strength $\sigma(a)\in[0,1]$ for every argument. Following the standard modular view of these semantics \cite{BaroniRagoToni2019Spectrum,Rago2016DFQuAD,Kampik2024Contribution}, a fixed aggregation map $\alpha$ combines child strengths and an influence map $\iota$ updates the parent strength:
\begin{equation}
\label{eq:method_modular_update}
  \sigma(a)
  =
  \iota_{w(a)}
  \left(
    \alpha_{\pi(a)}
    \bigl(\sigma(c_1),\ldots,\sigma(c_n)\bigr)
  \right).
\end{equation}
For leaves, this update reduces to $\sigma(a)=w(a)$. We use the DF-QuAD semantics in our implementation~\cite{Rago2016DFQuAD}.

\noindent
\textbf{Problem Setup.}
Building on the document-level extraction task defined in Section~\ref{subsec:def-task}, let $d\in\Dcal$ denote a \cti report and let $E=\{e_1,\ldots,e_K\}$ be the set of $K$ experts that jointly produce the predicted entity set $\widehat{Y}_{\Bcal}(d)$ over the benchmark label space $\Ycal_{\Bcal}$.
 
\subsection{Two-Phase Document-Level Extraction}
\label{subsec:method_two_phase}
 
\method separates extraction into two separate phases. The first phase is dedicated to candidate generation by constructing a union set of proposed candidates from each of the participating experts from the full document. The second phase is a verification step where each of the disputed candidates from the set undergoes a structured discussion between the experts, which in turn generates a QBAF to be used for the final decision.
 
\noindent
\textbf{Phase 1: candidate union generation.}
Each expert independently reads the full report and produces a structured response $g_e(d)$ containing candidate MITRE entities. We parse each response into a normalized candidate set
\begin{equation}
\label{eq:method_phase1_expert_set}
  C_e(d)
  =
  \Parse(g_e(d)),
  \qquad e\in E,
\end{equation}
where $\Parse(\cdot)$ extracts and normalizes MITRE identifiers from the model output to the benchmark label space $\Ycal_{\Bcal}$. The initial candidate pool is the union of all expert predictions:
\begin{equation}
\label{eq:method_candidate_union}
  C_{\Bcal}(d)
  =
  \bigcup_{e\in E} C_e(d).
\end{equation}
For each candidate $y\in C_{\Bcal}(d)$, we compute its vote count
\begin{equation}
\label{eq:method_vote_count}
  v_y(d)
  =
  \sum_{e\in E}\mathbb{I}[y\in C_e(d)],
\end{equation}
where $\mathbb{I}[\cdot]$ is the indicator function, equal to $1$ if its argument holds and $0$ otherwise. Candidates proposed by every expert are accepted directly:
\begin{equation}
\label{eq:method_unanimous_set}
  U_{\Bcal}(d)
  =
  \{y\in C_{\Bcal}(d): v_y(d)=K\}.
\end{equation}
The remaining candidates are disputed candidates:
\begin{equation}
\label{eq:method_debate_set}
  D_{\Bcal}(d)
  =
  C_{\Bcal}(d)\setminus U_{\Bcal}(d).
\end{equation}
This pruning is applied to reduce inference cost and reserve discussion for cases where expert disagreement makes verification informative.
 
\noindent
\textbf{Phase 2: per-candidate QBAF verification.}
For every disputed candidate $y\in D_{\Bcal}(d)$, \method constructs a separate verification discussion centered on the claim
\begin{equation}
\label{eq:method_fixed_root}
  m_y := \text{``MITRE entity } y \text{ is present in this document.''}
\end{equation}
This root argument is fixed by the system as a neutral, verifiable claim. The experts are prompted to generate support and attack arguments grounded in the report. To coordinate these outputs into a well-formed QBAF, \method employs a dedicated \textit{orchestrator} LLM that acts as an overseer of the verification process: it extracts and assigns individual arguments into a support or attack polarity with respect to $m_y$, and scores non-root arguments according to a set criterion. This yields a candidate-specific QBAF
\begin{equation}
\label{eq:method_per_entity_qbaf}
  Q_y
  =
  \langle \Acal_y, R_y^{-}, R_y^{+}, w_y\rangle,
  \qquad m_y\in\Acal_y.
\end{equation}
Since $m_y$ is a fixed neutral claim rather than an argument generated by the experts, we set
\begin{equation}
\label{eq:method_neutral_root}
  w_y(m_y)=0.5.
\end{equation}
All other arguments receive orchestrator-assigned base scores. After evaluating $Q_y$ under Eq.~\eqref{eq:method_modular_update}, the candidate score is the root strength
\begin{equation}
\label{eq:method_candidate_score}
  s_y(d) := \sigma_y(m_y).
\end{equation}
Then, the final output combines unanimous candidates with disputed candidates whose verification score exceeds a threshold $\tau\in[0,1]$:
\begin{equation}
\label{eq:method_final_prediction}
  \widehat{Y}_{\tau,\Bcal}(d)
  =
  U_{\Bcal}(d)
  \cup
  \{y\in D_{\Bcal}(d): s_y(d)\geq\tau\}.
\end{equation}
The threshold $\tau$ controls the operating point for precision~\ctag{C1}. Algorithm~\ref{alg:method_argora_cti} provides a complete step-by-step summary of the full extraction procedure.
 
\begin{algorithm}[t]
\caption{Document-Level \attack Extraction with \method}
\label{alg:method_argora_cti}
\begin{algorithmic}[1]
\REQUIRE Document $d$, experts $E=\{e_1,\ldots,e_K\}$, threshold $\tau$
\ENSURE Predicted normalized entity set $\widehat{Y}_{\tau,\Bcal}(d)$
\STATE $C_{\Bcal}(d)\leftarrow\emptyset$
\FOR{each expert $e\in E$}
    \STATE Generate full-document response $g_e(d)$
    \STATE $C_e(d)\leftarrow \Parse(g_e(d))$
    \STATE $C_{\Bcal}(d)\leftarrow C_{\Bcal}(d)\cup C_e(d)$
\ENDFOR
\FOR{each candidate $y\in C_{\Bcal}(d)$}
    \STATE $v_y(d)\leftarrow \sum_{e\in E}\mathbb{I}[y\in C_e(d)]$
\ENDFOR
\STATE $U_{\Bcal}(d)\leftarrow \{y\in C_{\Bcal}(d):v_y(d)=K\}$
\STATE $D_{\Bcal}(d)\leftarrow C_{\Bcal}(d)\setminus U_{\Bcal}(d)$
\STATE $\widehat{Y}_{\tau,\Bcal}(d)\leftarrow U_{\Bcal}(d)$
\FOR{each candidate $y\in D_{\Bcal}(d)$}
    \STATE Fix root claim $m_y$: entity $y$ is present in $d$
    \STATE Build $Q_y=\langle\Acal_y,R_y^{-},R_y^{+},w_y\rangle$ with $w_y(m_y)=0.5$
    \STATE Evaluate $Q_y$ and set $s_y(d)\leftarrow\sigma_y(m_y)$
    \IF{$s_y(d)\geq\tau$}
        \STATE $\widehat{Y}_{\tau,\Bcal}(d)\leftarrow\widehat{Y}_{\tau,\Bcal}(d)\cup\{y\}$
    \ENDIF
\ENDFOR
\RETURN $\widehat{Y}_{\tau,\Bcal}(d)$
\end{algorithmic}
\end{algorithm}

\subsection{Reasoning-Based Expert Construction}
\label{subsec:method_reasoning_experts}
 
\method requires experts that can both propose candidates and justify them during verification. Generic LLM justifications, however, may be shallow or poorly aligned with the evidential requirements of \attack mapping \ctag{C2}. We therefore construct supervised fine-tuning corpora containing evidence-grounded rationales, following prior work on reasoning supervision
~\cite{ouyang2022training_instructgpt,zelikman2022star,
hsieh2023distilling_step_by_step,lightman2023letsverify}. \method utilizes \textbf{three} experts, each targeting a failure mode of naive document-level extraction: (i) accepting plausible techniques without direct textual support; (ii) confusing semantically adjacent techniques; and
(iii) missing techniques whose evidence is distributed across multiple non-contiguous passages.

The public training splits of TRAM2 and AnnoCTR provide manually annotated
gold labels but no detailed explanations. We therefore use a teacher model
($M_{\mathrm{teach}}=\texttt{GPT-5.4-mini}$) to generate rationales
conditioned on each gold entity. For each training context
$x\in\Xcal_{\Bcal}^{\mathrm{train}}$ and gold entity
$y\in Y_{\Bcal}(x)$,
\begin{equation}
\label{eq:method_teacher_rationale}
  r_y(x)
  =
  M_{\mathrm{teach}}(x,y;\Pi_{\mathrm{just}}).
\end{equation}
The teacher is prompted to justify the entity using visible report evidence.
Web search is enabled to retrieve current \attack definitions and behavioral
descriptions when generating these rationales~\cite{nakano2021webgpt}.

\noindent
\textbf{Reasoning-Based Experts.}
Each expert uses a distinct response format corresponding
to one of the failure modes above.
 
The \emph{Evidence-Anchored Grounding Expert} ($e_{\mathrm{anchor}}$)
is motivated by extractive rationale methods that ground predictions in
input spans~\cite{lei-etal-2016-rationalizing}. It confirms a technique only when an exact document phrase can be quoted as an anchor, followed by a short justification; rejected techniques explicitly state that no direct textual anchor exists.
 
The \emph{Contrastive Evidence Expert} ($e_{\mathrm{contrast}}$) builds on
contrastive explanation, where a prediction is justified relative to plausible alternatives~\cite{jacovi-etal-2021-contrastive}. In addition to a
verbatim anchor, it explicitly rules out a similar
alternative and explains why the evidence does not support it. These alternatives are sampled from sibling techniques representing different mechanisms.
 
The \emph{Attack-Flow Inference Expert} ($e_{\mathrm{inference}}$) is
motivated by prior CTI work that reconstructs causal and temporal attack
behavior from threat reports~\cite{satvat2021extractor}. It first reconstructs
the attack progression in chronological and causal order without technique labels, describing attacker actions, affected resources, and dependencies. Gold techniques are then mapped to the reconstructed phases with explicit justification.

The three experts jointly form
\begin{equation}
\label{eq:method_expert_set}
  E_{\mathrm{FT}}
  =
  \{e_{\mathrm{anchor}},e_{\mathrm{contrast}},e_{\mathrm{inference}}\}.
\end{equation}
Their complementary evidential perspectives provide epistemic diversity during verification: support and attack arguments for the same candidate arise from distinct reasoning structures rather than near-duplicate reasoning styles. We empirically evaluate this property in our experiments.

\noindent
\textbf{Backbone selection.}
Expert backbones are selected per benchmark (TRAM2 and AnnoCTR) through a pilot single-pass evaluation on the training split.
GPT-4.1-mini performs consistently better on AnnoCTR, so all three experts
use it~($\ddagger$). On TRAM2, GPT-4o-mini and GPT-4.1-mini are both
competitive; the best-performing pool uses one GPT-4.1-mini and two
GPT-4o-mini experts~($\dagger$).

\subsection{Adversarial CTI Perturbations}
\label{subsec:method_adversarial}
 
To evaluate robustness beyond clean benchmark data, we construct three adversarial variants of the TRAM2 evaluation split, one for each of the adversary capability classes from Section~\ref{subsec:def-threatmodel} \ctag{C3}. The gold label set is preserved in all three cases, so performance changes measure extractor fragility to text-level manipulation.
 
\noindent
\textbf{Conflicting TTP statement injection:}
Insert plausible analyst-style statements that mention the same malware, actor, or campaign context but attribute behavior to techniques not present in the gold label set. This tests whether a model accepts plausible but unsupported technique claims.

\noindent
\textbf{Cross-report contamination:}
Interleave unlabeled sentences sampled from unrelated TRAM2 reports. This tests whether an extractor is distracted by contaminated \cti context.
 
\noindent
\textbf{Targeted semantic poisoning:}
Rewrite labeled sentences to preserve the same malicious behavior while weakening entity-signaling keywords, tool names, or phrases. This tests whether the model can recover techniques from behavioral meaning rather than lexical cues alone.

The implementation for the three adversarial perturbations is similar to that of Section~\ref{subsec:method_reasoning_experts}, where we use a teacher model to generate these perturbations.
\section{Evaluation: MITRE Entity Extraction}
\label{sec:evaluation_main}

We evaluate \method{} on two public \cti extraction benchmarks.
Our evaluation addresses six research questions:
\textbf{RQ1 (Effectiveness):} Does \method{} improve document-level extraction over sentence-level and frontier-LLM baselines?
\textbf{RQ2 (Reasoning supervision):} Does reasoning-based fine-tuning improve multi-expert verification beyond answer-only fine-tuning, even when it does not improve standalone extraction?
\textbf{RQ3 (Existing systems):} How does \method{} compare with recent \cti extraction systems under a common \attack output space and comparable backbones?
\textbf{RQ4 (Cost analysis):} What runtime and API-cost overhead does multi-expert verification incur?
\textbf{RQ5 (Score discriminability):} Does reasoning supervision help QBAF verification distinguish supported from unsupported candidates?
\textbf{RQ6 (Verification mechanism):} Is structured multi-expert QBAF verification more effective than a single-call LLM verifier?

\subsection{Experimental Setup}
\label{subsec:eval_main_setup}

\noindent\textbf{Datasets and metrics.}
We evaluate on TRAM2 and AnnoCTR.
TRAM2 provides technique-level annotations for \cti reports~\cite{tram}; following the split used by the SoK~\cite{sok_ttp_extraction_2025}, we evaluate on 31 held-out reports. As the original TRAM2 dataset is sentence-level, we convert the dataset to document-level by merging all sentences from the same document to a single datapoint and taking the union of sentence-level annotations as the gold label.
AnnoCTR contains techniques, tactics, threat groups, and software~\cite{annoctr}; we use the corresponding SoK split of 34 held-out documents and normalize sub-techniques to their parent techniques for scoring. Additional dataset statistics are provided in Appendix~\ref{app:data-statistics}.
We report document-level macro-averaged precision, recall, and F1 using exact normalized \attack identifier matches.

\noindent\textbf{Baselines and configurations.}
Our primary comparison (Table~\ref{tab:main_results_summary}) includes:
(i) the strongest sentence-level NER pipeline from the unified SoK evaluation~\cite{sok_ttp_extraction_2025};
(ii) single-pass full-document GPT-4o-mini, GPT-4.1-mini, GPT-5.4, and GPT-5.5 baselines~\cite{openai_models};
(iii) standalone answer-only ($e_{\mathrm{basic}}$) and reasoning-specialized ($e_{\mathrm{anchor}},e_{\mathrm{contrast}},e_{\mathrm{inference}}$) fine-tuned experts; and
(iv) \method{} with No-FT (3 non-finetuned LLM backbone experts), Basic-FT (3 $e_{\mathrm{basic}}$ experts), and Reasoning-FT ($e_{\mathrm{anchor}},e_{\mathrm{contrast}},e_{\mathrm{inference}}$) expert pools.
As described in Section~\ref{subsec:method_reasoning_experts}, the Reasoning-FT pool uses one GPT-4.1-mini and two GPT-4o-mini backbones on TRAM2, and three GPT-4.1-mini backbones on AnnoCTR. All QBAFs use DF-QuAD semantics~\cite{Rago2016DFQuAD}, all LLM calls use temperature~0, and the default operating threshold is $\tau=0.5$.
Across five repetitions of \method{} Reasoning FT on TRAM2, F1 is $0.649\pm0.015$ (95\% CI: $[0.631,0.668]$).

\subsection{Extraction Performance}
\label{subsec:eval_overall}
\label{subsec:eval_tram2}
\label{subsec:eval_annoctr}

\begin{table*}[t]
\centering
\caption{Document-level extraction results on TRAM2 and AnnoCTR
($^{\dagger}$one GPT-4.1-mini and two GPT-4o-mini experts;
$^{\ddagger}$three GPT-4.1-mini experts).
GPT-5.4 and GPT-5.5 are document-level baselines only (not used as \method{} experts).
Deltas (in pp -- percentage points) below each \method{} row are relative to the corresponding standalone
expert: GPT-4o-mini for No-FT, $e_{\mathrm{basic}}$ for Basic FT, and the
best-F1 reasoning expert ($e_{\mathrm{inference}}$) for Reasoning FT.
Bold: best per column.}
\label{tab:main_results_summary}
\scriptsize
\setlength{\tabcolsep}{2.8pt}
\renewcommand{\arraystretch}{1.12}
\resizebox{0.85\linewidth}{!}{%
\begin{tabular}{@{}p{2.85cm} p{3.15cm} *{6}{>{\centering\arraybackslash}p{0.95cm}}@{}}
\toprule
\multirow{2}{*}{\textbf{System}}
& \multirow{2}{*}{\textbf{Setting}}
& \multicolumn{3}{c}{\textbf{TRAM2}$^{\dagger}$}
& \multicolumn{3}{c}{\textbf{AnnoCTR}$^{\ddagger}$} \\
\cmidrule(lr){3-5}
\cmidrule(l){6-8}
&
& \textbf{P} & \textbf{R} & \textbf{F1}
& \textbf{P} & \textbf{R} & \textbf{F1} \\
\midrule
\multicolumn{8}{@{}l}{\textit{Sentence-level baseline}} \\
Full NER pipeline \cite{sok_ttp_extraction_2025}
& Sentence-level extraction
& 0.654 & 0.578 & 0.614
& 0.435 & 0.709 & 0.539 \\
\midrule
\multicolumn{8}{@{}l}{\textit{Document-level LLM baselines}} \\
GPT-4o-mini
& Document-level extraction
& 0.317 & 0.322 & 0.299
& 0.293 & 0.256 & 0.263 \\
GPT-4.1-mini
& Document-level extraction
& 0.292 & 0.392 & 0.319
& 0.340 & 0.432 & 0.362 \\
GPT-5.4
& Document-level extraction
& 0.341 & 0.510 & 0.401
& 0.575 & 0.653 & 0.598 \\
GPT-5.5
& Document-level extraction
& 0.335 & 0.561 & 0.411
& 0.528 & \textbf{0.743} & \textbf{0.603} \\
\midrule
\multicolumn{8}{@{}l}{\textit{Single finetuned expert instances}} \\
$e_{\mathrm{basic}}$
& Basic FT
& 0.676 & 0.530 & 0.577
& 0.541 & 0.515 & 0.512 \\
$e_{\mathrm{anchor}}$
& Reasoning FT
& 0.605 & 0.500 & 0.503
& 0.535 & 0.336 & 0.396 \\
$e_{\mathrm{contrast}}$
& Reasoning FT
& 0.573 & 0.552 & 0.534
& 0.580 & 0.300 & 0.379 \\
$e_{\mathrm{inference}}$
& Reasoning FT
& 0.662 & 0.492 & 0.546
& 0.490 & 0.398 & 0.424 \\
\midrule
\multicolumn{8}{@{}l}{%
  \begin{tabular}[t]{@{}l@{}}
  \textit{Multi-expert orchestration (ours)}\\[-0.15em]
  {\scriptsize Expert pool: GPT-4o-mini / GPT-4.1-mini}
  \end{tabular}%
} \\
\method
& No-FT (3$\times$ GPT-4o-mini)
& 0.348 & 0.342 & 0.327
& 0.350 & 0.449 & 0.357 \\[-3pt]
& & {\scriptsize\textcolor{green!60!black}{+3.1\,pp}}
  & {\scriptsize\textcolor{green!60!black}{+2.0\,pp}}
  & {\scriptsize\textcolor{green!60!black}{+2.8\,pp}}
  & {\scriptsize\textcolor{green!60!black}{+5.7\,pp}}
  & {\scriptsize\textcolor{green!60!black}{+19.3\,pp}}
  & {\scriptsize\textcolor{green!60!black}{+9.4\,pp}} \\[2pt]
\method
& Basic FT
& 0.674 & 0.617 & 0.619
& 0.565 & 0.483 & 0.502 \\[-3pt]
& & {\scriptsize\textcolor{red!70!black}{$-$0.2\,pp}}
  & {\scriptsize\textcolor{green!60!black}{+8.7\,pp}}
  & {\scriptsize\textcolor{green!60!black}{+4.2\,pp}}
  & {\scriptsize\textcolor{green!60!black}{+2.4\,pp}}
  & {\scriptsize\textcolor{red!70!black}{$-$3.2\,pp}}
  & {\scriptsize\textcolor{red!70!black}{$-$1.0\,pp}} \\[2pt]
\method
& Reasoning FT
& \textbf{0.722} & \textbf{0.635} & \textbf{0.649}
& \textbf{0.593} & 0.477 & 0.516 \\[-3pt]
& & {\scriptsize\textcolor{green!60!black}{+6.0\,pp}}
  & {\scriptsize\textcolor{green!60!black}{+14.3\,pp}}
  & {\scriptsize\textcolor{green!60!black}{+10.3\,pp}}
  & {\scriptsize\textcolor{green!60!black}{+10.3\,pp}}
  & {\scriptsize\textcolor{green!60!black}{+7.9\,pp}}
  & {\scriptsize\textcolor{green!60!black}{+9.2\,pp}} \\
\bottomrule
\end{tabular}%
}
\end{table*}

\noindent\textbf{RQ1: Overall effectiveness.}
Table~\ref{tab:main_results_summary} shows that stronger single-pass frontier models improve recall on TRAM2 but do not resolve its precision deficit: GPT-5.5 reaches $R=0.561$ but only $P=0.335$, compared with $P=0.654$ for the sentence-level NER pipeline.
\method{} (Reasoning FT) reaches $P/R/F1=0.722/0.635/0.649$, surpassing the NER pipeline on all three metrics and GPT-5.5 by 23.8~pp F1. The results also suggest that per-candidate verification preserves a substantially stronger precision-oriented operating point \ctag{C1}.

AnnoCTR presents a complementary transfer setting.
Here, \method{} (Reasoning FT) reaches the highest precision ($P=0.593$), although GPT-5.5 retains higher recall and F1.
This gap is consistent with the benchmark's training--test label mismatch: 15.5\% of techniques, 26.1\% of software, and 53.8\% of threat groups present in the test set are \textit{never} seen during fine-tuning, whereas TRAM2's test labels are fully covered by the training (fine-tuning) set.
Consequently, fine-tuned experts face a recall ceiling for unseen entities.

\noindent\textbf{RQ2: Benefits of reasoning supervision.}
The benefit of reasoning supervision emerges most clearly \emph{inside} \method{} rather than from standalone extraction.
On TRAM2, Reasoning FT improves F1 by 10.3~pp over the strongest standalone reasoning expert, compared with 4.2~pp for Basic FT over $e_{\mathrm{basic}}$.
On AnnoCTR, the same comparison yields $+9.2$~pp for Reasoning FT, whereas Basic FT decreases F1 by 1.0~pp.
Thus, reasoning-based experts need not be stronger standalone classifiers to be more useful participants in evidence-grounded QBAF verification \ctag{C2}.

\begin{figure}[ht!]
\centering
\includegraphics[width=0.92\columnwidth]{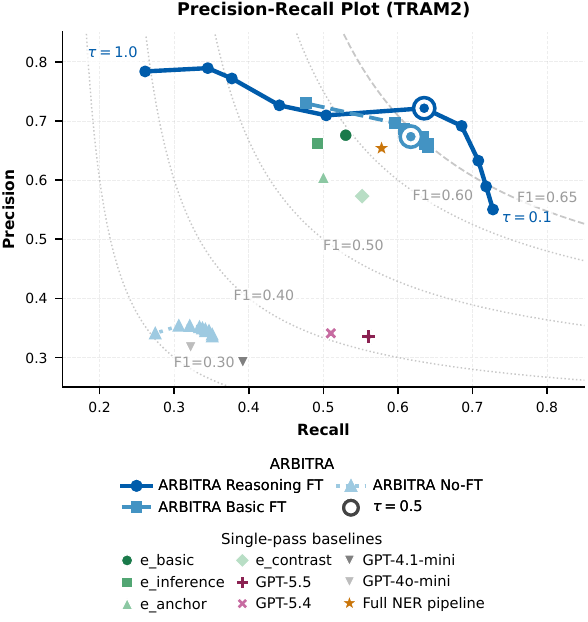}
\caption{TRAM2 precision--recall curve as $\tau$ varies from 1.0 to 0.1. Fixed points denote single-pass baselines. ISO-F1 contours use aggregate precision/recall, whereas Table~\ref{tab:main_results_summary} reports document-level macro-F1, so the values need not coincide.}
\label{fig:pr-curve-tram2}
\end{figure}

Figure~\ref{fig:pr-curve-tram2} further shows that $\tau$ provides explicit precision--recall tradeoff control: Reasoning FT traces the strongest frontier among the \method{} configurations, moving from high-precision unanimous acceptance toward higher recall as $\tau$ decreases.
Appendix~\ref{app:abl-exp-combo} further evaluates expert-pool composition.

\begin{table*}[t]
\centering
\caption{Document-level \textit{technique-only} extraction against recent \cti systems and LLMs. CTINexus uses our \attack-mapping bridge; TTPrint is reimplemented from its published prompts. We report matched-backbone, headline-backbone, and best-performing Reasoning-FT variants ($e_{\mathrm{anchor}}$, $e_{\mathrm{contrast}}$, $e_{\mathrm{inference}}$) for each applicable system. Bold: best per column among existing \cti systems and \method{}; italics: document-level LLM scores exceeding \method{}. Gray cells represent scores from the best-performing backbone variant for that specific system.}
\label{tab:baseline_comparison}
\scriptsize
\setlength{\tabcolsep}{2.8pt}
\renewcommand{\arraystretch}{1.12}
\resizebox{0.9\linewidth}{!}{%
\begin{tabular}{@{}p{3.4cm} p{3.0cm} *{6}{>{\centering\arraybackslash}p{0.95cm}}@{}}
\toprule
\multirow{2}{*}{\textbf{System}}
& \multirow{2}{*}{\textbf{Backbone}}
& \multicolumn{3}{c}{\textbf{TRAM2}}
& \multicolumn{3}{c}{\textbf{AnnoCTR (techniques only)}} \\
\cmidrule(lr){3-5}
\cmidrule(l){6-8}
&
& \textbf{P} & \textbf{R} & \textbf{F1}
& \textbf{P} & \textbf{R} & \textbf{F1} \\
\midrule
\multicolumn{8}{@{}l}{\textit{Document-level LLM baselines}} \\
Single-pass prompting & GPT-4o-mini
& 0.317 & 0.322 & 0.299
& 0.501 & 0.302 & 0.354 \\
Single-pass prompting & GPT-5.5
& \cellcolor{gray!15}0.335 & \cellcolor{gray!15}0.561 & \cellcolor{gray!15}0.411
& \cellcolor{gray!15}0.648 & \cellcolor{gray!15}\textit{0.651} & \cellcolor{gray!15}\textit{0.635} \\
\midrule
\multicolumn{8}{@{}l}{\textit{Existing \cti extraction systems}} \\
TTPXHunter \cite{ttpxhunter} & SecureBERT (finetuned)
& \cellcolor{gray!15}0.245 & \cellcolor{gray!15}0.531 & \cellcolor{gray!15}0.328
& \cellcolor{gray!15}0.220 & \cellcolor{gray!15}0.356 & \cellcolor{gray!15}0.265 \\
\addlinespace[2.5pt]
CTINexus \cite{ctinexus} +\,bridge & GPT-4o-mini
& \cellcolor{gray!15}0.355 & \cellcolor{gray!15}0.211 & \cellcolor{gray!15}0.246
& 0.307 & 0.213 & 0.229 \\
CTINexus \cite{ctinexus} +\,bridge & GPT-4.1
& 0.282 & 0.141 & 0.168
& 0.298 & 0.195 & 0.195 \\
CTINexus \cite{ctinexus} +\,bridge & $e_{\mathrm{contrast}}^{\mathrm{(TRAM2)}}$\,/\,$e_{\mathrm{contrast}}^{\mathrm{(AnnoCTR)}}$
& 0.326 & 0.196 & 0.228
& \cellcolor{gray!15}0.364 & \cellcolor{gray!15}0.222 & \cellcolor{gray!15}0.256 \\
\addlinespace[2.5pt]
TTPrint \cite{ttprint} & GPT-4o-mini
& 0.466 & 0.476 & 0.447
& 0.602 & 0.368 & 0.426 \\
TTPrint \cite{ttprint} & GPT-4o
& 0.484 & \cellcolor{gray!15}0.582 & \cellcolor{gray!15}0.494
& 0.521 & \cellcolor{gray!15}0.442 & 0.447 \\
TTPrint \cite{ttprint} & $e_{\mathrm{anchor}}^{\mathrm{(TRAM2)}}$\,/\,$e_{\mathrm{contrast}}^{\mathrm{(AnnoCTR)}}$
& \cellcolor{gray!15}0.574 & 0.278 & 0.338
& \cellcolor{gray!15}0.690 & 0.400 & \cellcolor{gray!15}0.487 \\
\midrule
\multicolumn{8}{@{}l}{\textit{Multi-expert orchestration (ours)}} \\
\rowcolor{blue!7}
\method{} (Reasoning FT) & $e_{\mathrm{anchor}}$ + $e_{\mathrm{contrast}}$ + $e_{\mathrm{inference}}$
& \textbf{0.722} & \textbf{0.635} & \textbf{0.649}
& \textbf{0.711} & \textbf{0.476} & \textbf{0.554} \\
\bottomrule
\end{tabular}%
}
\end{table*}

\subsection{Comparison with Existing CTI Systems}
\label{subsec:eval_external}

In order to answer RQ3, we next compare \method{} against three recent \cti-specific extraction systems that represent complementary designs: TTPXHunter~\cite{ttpxhunter}, a supervised sentence-level TTP classifier; CTINexus~\cite{ctinexus}, an LLM-based cybersecurity knowledge-graph construction pipeline; and TTPrint~\cite{ttprint}, a document-level LLM pipeline that separates technique proposal from evidence-grounded verification.
Because these systems expose different output spaces, we evaluate their largest directly comparable target: \attack techniques.
TTPXHunter and TTPrint are designed around \attack technique prediction rather than the broader tactic, group, and software labels included in AnnoCTR.
CTINexus supports broader cybersecurity ontologies, but produces entity--relation graphs rather than \attack identifiers, so we add a fixed LLM \emph{bridge} that maps its graph output to target techniques.

For backbone fairness, we evaluate CTINexus with GPT-4.1 as a cost-efficient substitute for GPT-4, its strongest reported backbone, and with GPT-4o-mini for a matched-backbone setting.
As TTPrint does not provide a public implementation at the time of our evaluation, we perform a best-effort reimplementation following its described pipeline and published prompt templates using GPT-4o, its best-performing reported backbone, and GPT-4o-mini. We additionally instantiate both LLM-based pipelines with our Reasoning-FT experts to test whether \method{}'s gains persist when competing architectures receive the same domain-specialized models.

\noindent\textbf{RQ3: Comparison with existing systems.}
The baseline comparison results are shown in Table~\ref{tab:baseline_comparison}. \method{} outperforms all three \cti systems on both datasets across all metrics.
On TRAM2, it reaches $F1=0.649$, 15.5~pp above the strongest TTPrint configuration ($0.494$).
On technique-only AnnoCTR, \method{} reaches $P/R/F1=0.711/0.476/0.554$; providing CTINexus and TTPrint with our Reasoning-FT experts does not close the gap, indicating that the improvement is not explained by domain fine-tuning alone.

\subsection{Cost Analysis}
\label{subsec:eval_cost}

\begin{table}[t]
\centering
\caption{Per-document median runtime and output-token API cost on the TRAM2 evaluation (over 31 documents).}
\label{tab:cost_analysis_cti}
\footnotesize
\renewcommand{\arraystretch}{1.12}
\resizebox{0.8\columnwidth}{!}{%
\begin{tabular}{l r r r}
\toprule
\textbf{System} & \textbf{Output tok.} & \textbf{Runtime} & \textbf{Output cost} \\
\midrule
GPT-5.4 & 2{,}865 & 43.6s & \$0.0430 \\
GPT-5.5 & 2{,}534 & 41.1s & \$0.0760 \\
\textbf{\method} & \textbf{26{,}513} & \textbf{146.0s} & \textbf{\$0.0214} \\
\bottomrule
\end{tabular}%
}
\vspace{0.25em}

\footnotesize\emph{Note.} The \method total includes three Reasoning-FT expert calls and the orchestrator. Using raw per-token billing rates, this puts \method at approximately 50\% of GPT-5.4 and 28\% of GPT-5.5's per-document output cost despite producing more output tokens (as of August 2026).
\end{table}

\noindent\textbf{RQ4: Cost analysis.}
Table~\ref{tab:cost_analysis_cti} reports per-document runtime and output-token API cost on TRAM2.
We use output-token rather than total token cost because \method{} repeatedly reuses long report and instruction prefixes across candidate discussions; shared input prefixes can benefit from API prompt caching, making input cost dependent on cache-hit behavior and deployment details~\cite{openai_models}.
Output tokens are unaffected by prompt-prefix caching and therefore provide a more stable comparison of marginal generation cost.

Although \method{} produces substantially more output through multi-expert verification, its median output cost is \$0.0214 per report, approximately 50\% of GPT-5.4 (\$0.0430) and 28\% of GPT-5.5 (\$0.0760), because its expert backbones have substantially lower per-token rates. The trade-off is latency: \method{} requires 146.0s per report versus 41--44s for the frontier single-pass models, largely due to per-candidate discussions.
For batch-oriented \cti extraction, this yields a practical trade-off of increased latency for substantially lower frontier-model output cost and stronger precision-oriented extraction performance.

\begin{table*}[t]
\centering
\caption{%
  Ablation of the Phase~2 verification mechanism on TRAM2 and AnnoCTR:
  per-candidate QBAF Discussion (\method{}) vs.\ a single Verifier LLM.
  Conditions are detailed in the section headers; \method{} (QBAF) Reasoning~FT
  values in conditions~2 and~3 are identical and repeated for direct comparison.
  $\Delta$ rows show \method{} minus Verifier LLM in percentage points~(pp);
  \textcolor{green!60!black}{green}/\textcolor{red!70!black}{red}~=~QBAF
  advantage/disadvantage. \textbf{Bold}: higher value within each condition block.%
}
\label{tab:ablation_phase2}
\scriptsize
\setlength{\tabcolsep}{2.8pt}
\renewcommand{\arraystretch}{1.12}
\resizebox{0.85\linewidth}{!}{%
\begin{tabular}{@{}p{2.85cm} p{3.55cm}
  *{3}{>{\centering\arraybackslash}p{0.95cm}}
  *{3}{>{\centering\arraybackslash}p{0.95cm}}@{}}
\toprule
\multirow{2}{*}{\textbf{System}}
  & \multirow{2}{*}{\textbf{Verification Mechanism}}
  & \multicolumn{3}{c}{\textbf{TRAM2}}
  & \multicolumn{3}{c}{\textbf{AnnoCTR}} \\
\cmidrule(lr){3-5}\cmidrule(l){6-8}
  &
  & \textbf{P} & \textbf{R} & \textbf{F1}
  & \textbf{P} & \textbf{R} & \textbf{F1} \\
\midrule

\multicolumn{8}{@{}l}{%
  \textit{Condition 1: No fine-tuning\ \ (Phase~1: 3$\times$ GPT-4o-mini)}%
} \\[1pt]
\method{} (QBAF)
  & QBAF Discussion
  & \textbf{0.348} & 0.342 & 0.327
  & 0.350 & \textbf{0.449} & 0.357 \\
\method{} (Verifier LLM)
  & Verifier LLM (GPT-4o-mini)
  & 0.338 & \textbf{0.379} & \textbf{0.337}
  & \textbf{0.365} & 0.438 & \textbf{0.370} \\[-3pt]
  & {\scriptsize\textit{$\Delta$ QBAF $-$ Verifier}}
  & {\scriptsize\textcolor{green!60!black}{+1.0\,pp}}
    & {\scriptsize\textcolor{red!70!black}{$-$3.7\,pp}}
    & {\scriptsize\textcolor{red!70!black}{$-$1.0\,pp}}
  & {\scriptsize\textcolor{red!70!black}{$-$1.5\,pp}}
    & {\scriptsize\textcolor{green!60!black}{+1.1\,pp}}
    & {\scriptsize\textcolor{red!70!black}{$-$1.3\,pp}} \\[2pt]

\midrule

\multicolumn{8}{@{}l}{%
  \textit{Condition 2: Reasoning FT\ \ (Phase~1: $e_{\mathrm{anchor}}$,
  $e_{\mathrm{contrast}}$, $e_{\mathrm{inference}}$;\ verifier: GPT-4o-mini)}%
} \\[1pt]
\method{} (QBAF)
  & QBAF Discussion
  & \textbf{0.722} & \textbf{0.635} & \textbf{0.649}
  & \textbf{0.593} & 0.477 & \textbf{0.516} \\
\method{} (Verifier LLM)
  & Verifier LLM (GPT-4o-mini)
  & 0.714 & 0.460 & 0.540
  & 0.521 & \textbf{0.481} & 0.478 \\[-3pt]
  & {\scriptsize\textit{$\Delta$ QBAF $-$ Verifier}}
  & {\scriptsize\textcolor{green!60!black}{+0.8\,pp}}
    & {\scriptsize\textcolor{green!60!black}{+17.5\,pp}}
    & {\scriptsize\textcolor{green!60!black}{+10.9\,pp}}
  & {\scriptsize\textcolor{green!60!black}{+7.2\,pp}}
    & {\scriptsize\textcolor{red!70!black}{$-$0.4\,pp}}
    & {\scriptsize\textcolor{green!60!black}{+3.8\,pp}} \\[2pt]

\midrule

\multicolumn{8}{@{}l}{%
  \textit{Condition 3: Reasoning FT\ \ (Phase~1: $e_{\mathrm{anchor}}$,
  $e_{\mathrm{contrast}}$, $e_{\mathrm{inference}}$;\ verifier: $e_{\mathrm{best}}$)}%
} \\[1pt]
\method{} (QBAF)
  & QBAF Discussion
  & \textbf{0.722} & \textbf{0.635} & \textbf{0.649}
  & \textbf{0.593} & \textbf{0.477} & \textbf{0.516} \\
\method{} (Verifier LLM)
  & Verifier LLM ($e_{\mathrm{best}}$)
  & 0.720 & 0.422 & 0.503
  & 0.566 & 0.475 & 0.495 \\[-3pt]
  & {\scriptsize\textit{$\Delta$ QBAF $-$ Verifier}}
  & {\scriptsize\textcolor{green!60!black}{+0.2\,pp}}
    & {\scriptsize\textcolor{green!60!black}{+21.3\,pp}}
    & {\scriptsize\textcolor{green!60!black}{+14.6\,pp}}
  & {\scriptsize\textcolor{green!60!black}{+2.7\,pp}}
    & {\scriptsize\textcolor{green!60!black}{+0.2\,pp}}
    & {\scriptsize\textcolor{green!60!black}{+2.1\,pp}} \\

\bottomrule
\end{tabular}%
}
\end{table*}

\subsection{QBAF Score Discriminability Under Reasoning vs.\ Answer-Only Fine-Tuning}
\label{subsec:eval_ft_style}

\begin{figure}[!t]
  \centering
  \includegraphics[
    width=0.8\linewidth,
    keepaspectratio,
    clip
  ]{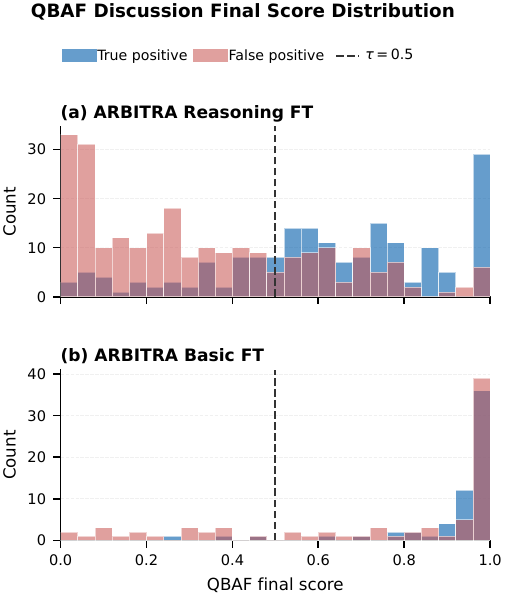}
  \caption{QBAF scores for disputed TRAM2 candidates under Reasoning~FT
    and Basic~FT. TP/FP denote candidates that do/do not match a gold
    label; the dashed line marks $\tau=0.5$.}
  \label{fig:qbaf-score}
\end{figure}

\noindent\textbf{RQ5 (Score discriminability):}
Having answered \textbf{RQ1--RQ4}, we examine why reasoning-specialized experts and QBAF verification are effective together. Figure~\ref{fig:qbaf-score} shows a clear difference between the reasoning-based and answer-only fine-tuning methods. With Reasoning~FT, gold-positive candidates
concentrate at higher QBAF scores while unsupported candidates shift
toward lower scores, making $\tau=0.5$ an effective filter. Under
Basic~FT, this separation largely collapses, with many unsupported
candidates receiving scores near 1. This is consistent with
reasoning-specialized experts providing evidence-specific support and
attack arguments that the orchestrator can score discriminatively,
whereas answer-only supervision provides weaker evidence for
verification. Thus, reasoning supervision is valuable not merely for
standalone extraction, but specifically for producing useful arguments
within \method{} \ctag{C2}.

\subsection{Expert QBAF vs.\ Single-Call Verification}
\label{subsec:eval_verifier}

\textbf{RQ6 (Verification mechanism):} In this section, we ask if structured multi-expert QBAF
verification is more effective than assigning each candidate to a single LLM verifier.

\begin{figure}[!htbp]
  \centering
  \includegraphics[
    width=0.8\linewidth,
    keepaspectratio,
    clip
  ]{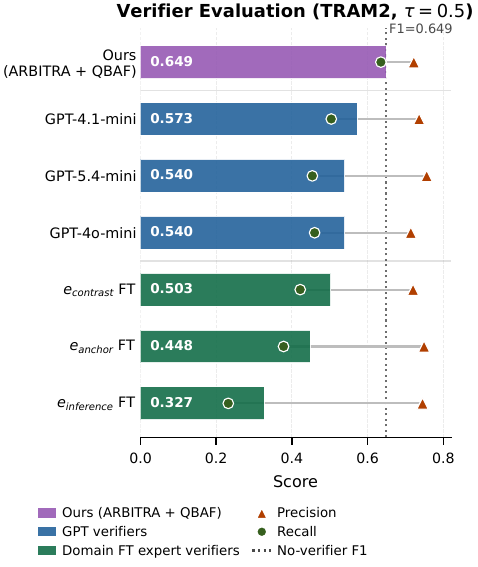}
  \caption{TRAM2 verification results at $\tau=0.5$ with the candidate union phase fixed. Only the verification phase varies between \method{}'s QBAF and single-call generic or domain-finetuned LLM verifiers.}
  \label{fig:verifier-eval-f1}
\end{figure}

This ablation also isolates an important architectural distinction from
TTPrint~\cite{ttprint}. TTPrint ultimately assigns each proposed
technique to a single LLM verifier conditioned on localized evidence
and its \attack definition, whereas \method{} adjudicates disputed
candidates through arguments from multiple experts. Although this ablation is not a reproduction of TTPrint's full pipeline (this comparison was already handled in Section~\ref{subsec:eval_external}), it tests a narrower scoped distinction by replacing only the QBAF verification phase with a single-call LLM-as-judge
baseline~\cite{zheng2023judging_llm}. We test three generic LLMs and the three Reasoning~FT experts as verifiers; Figure~\ref{fig:verifier-eval-f1}
reports TRAM2, while Table~\ref{tab:ablation_phase2} additionally
covers No-FT and AnnoCTR settings.

Every single-verifier configuration under Reasoning~FT falls below
\method{} ($F1=0.649$): the strongest, GPT-4.1-mini, reaches
$F1=0.573$, while domain-finetuned verifiers perform as low as
$0.327$. The gap is predominantly recall, indicating that single
verifiers reject supported candidates despite maintaining high
precision. Table~\ref{tab:ablation_phase2} further shows that this advantage largely disappears without fine-tuning, but becomes substantial with Reasoning~FT: against GPT-4o-mini verification,
QBAF gains 10.9~pp F1 on TRAM2 and 3.8~pp on AnnoCTR, and gains
14.6/2.1~pp when compared with the best reasoning-expert verifier.
Together with RQ5, these results show that \method{} benefits from the
\emph{combination} of evidence-rich expert reasoning and structured
multi-expert verification, rather than either component alone
\ctag{C2}.
\section{Evaluation: Adversarial CTI Robustness}
\label{sec:evaluation_adversarial}

We next evaluate \method under the adversarial CTI threat model defined in
Section~\ref{subsec:def-threatmodel} \ctag{C3}. Recall that the three
perturbation classes target two distinct degradation objectives:
\textit{conflicting TTP statements} and \textit{cross-report contamination}
aim to inject false positives (\fpobj), whereas \textit{semantic poisoning}
aims to suppress true positives (\tpobj). We therefore analyze the attacks
according to their intended failure mode: precision is the primary robustness measure for the two \fpobj{} variants and recall for the \tpobj{} variant.

\noindent
\textbf{Setup.}
We evaluate all systems on the three adversarial TRAM2 variants constructed in
Section~\ref{subsec:method_adversarial}, alongside their clean TRAM2 results.
The original gold label sets are unchanged by the perturbations, such that
performance changes reflect sensitivity to the modified report text. Consistent
with Section~\ref{sec:evaluation_main}, Table~\ref{tab:adv_summary} reports
document-level macro-averaged precision and recall. We additionally report the mean relative precision change from each system's own clean baseline (Mean~$\Delta$P) as an aggregate measure of precision stability across perturbations.

\subsection{Conflicting TTP Statements}
\label{subsec:adv_conflicting}

Conflicting TTP statements directly test the \fpobj{} objective by introducing
plausible but unsupported technique attributions. Accordingly, the principal
signal is whether these injected claims reduce extraction precision.
\method retains precision of $0.662$, compared with $0.722$ on clean TRAM2,
corresponding to an $8.3\%$ relative decrease. In contrast, GPT-5.4 falls from
$0.341$ to $0.284$ ($-16.7\%$) and GPT-5.5 from $0.335$ to $0.266$
($-20.6\%$). Thus, although the injected statements affect all systems,
\method preserves a substantially stronger precision-oriented operating point
and retains the highest absolute precision on the perturbed reports.

This robustness does not arise simply from rejecting more candidates:
\method also achieves the highest recall on this variant ($0.576$).
The result is consistent with the intended role of verification as a plausible technique mention is insufficient by itself when a disputed candidate must accumulate evidence-grounded support through multi-expert argumentation before acceptance.

\begin{table*}[t]
\centering
\caption{
Robustness summary on clean TRAM2 and three adversarial variants.
Cells report precision/recall (P/R).
Adversarial variants are annotated by their primary degradation objective:
\fpobj{} denotes false-positive injection and
\tpobj{} denotes true-positive suppression.
Mean $\Delta$P reports the average relative precision change from each
model's own clean baseline across the three variants
(negative values indicate degradation).
Shaded cells range from white (least degradation) to red (most degradation).
}
\label{tab:adv_summary}

\footnotesize
\renewcommand{\arraystretch}{1.12}

\resizebox{0.95\linewidth}{!}{%
\begin{tabular}{p{5.5cm} c c c c c}
\toprule
\textbf{Model}
&
\textbf{Clean}
&
\textbf{Conflicting}~(\fpobj)
&
\textbf{Cross-report}~(\fpobj)
&
\textbf{Semantic}~(\tpobj)
&
\textbf{Mean $\Delta$P}
\\

&
\textbf{P/R}
&
\textbf{P/R}
&
\textbf{P/R}
&
\textbf{P/R}
&
\\

\midrule

GPT-4o-mini (CoT)
& 0.317 / 0.322
& 0.289 / 0.302
& 0.284 / 0.284
& 0.285 / 0.251
& \cellcolor{red!21!white}$-9.8\%$
\\

GPT-5.4
& 0.341 / 0.510
& 0.284 / 0.461
& 0.286 / 0.446
& 0.328 / 0.457
& \cellcolor{red!35!white}$-12.2\%$
\\

GPT-5.5
& 0.335 / 0.561
& 0.266 / 0.512
& 0.278 / 0.504
& 0.320 / 0.511
& \cellcolor{red!45!white}$-14.0\%$
\\

\midrule

$e_{\mathrm{basic}}$ (finetuned GPT-4o-mini)
& 0.676 / 0.530
& 0.600 / 0.557
& 0.596 / 0.533
& 0.655 / 0.510
& \cellcolor{red!15!white}$-8.7\%$
\\

$e_{\mathrm{anchor}}$ (finetuned GPT-4.1-mini)
& 0.605 / 0.500
& 0.543 / 0.503
& 0.502 / 0.517
& 0.633 / 0.495
& \cellcolor{red!8!white}$-7.5\%$
\\

$e_{\mathrm{contrast}}$ (finetuned GPT-4o-mini)
& 0.573 / 0.552
& 0.537 / 0.480
& 0.513 / 0.470
& 0.536 / 0.473
& \cellcolor{red!9!white}$-7.7\%$
\\

$e_{\mathrm{inference}}$ (finetuned GPT-4o-mini)
& 0.662 / 0.492
& 0.606 / 0.474
& 0.561 / 0.475
& 0.613 / 0.469
& \cellcolor{red!24!white}$-10.4\%$
\\

\midrule

\method{} ($e_{\mathrm{anchor}}$, $e_{\mathrm{contrast}}$,
$e_{\mathrm{inference}}$)
& \textbf{0.722 / 0.635}
& \textbf{0.662 / 0.576}
& \textbf{0.657 / 0.551}
& \textbf{0.696 / 0.545}
& \cellcolor{red!5!white}$\mathbf{-7.0\%}$
\\

\bottomrule
\end{tabular}%
}
\end{table*}

\subsection{Cross-Report Contamination}
\label{subsec:adv_crossreport}

Cross-report contamination provides a complementary \fpobj{} test in which the
misleading content is itself legitimate CTI, but originates from an unrelated
report. \method again retains the highest precision, decreasing from $0.722$
to $0.657$ ($-9.0\%$ relative). The corresponding losses are substantially
larger for GPT-5.4 ($-16.1\%$) and GPT-5.5 ($-17.0\%$). At the same time,
\method maintains recall of $0.551$, also the highest among the evaluated
systems on this variant.

The standalone $e_{\mathrm{anchor}}$ expert illustrates why this perturbation
is particularly challenging for evidence-localized extraction. Its recall
increases slightly from $0.500$ to $0.517$, while its precision falls from
$0.605$ to $0.502$ ($-17.0\%$ relative). This pattern is consistent with
contaminating passages providing genuine-looking textual anchors that surface
additional candidates but are unsupported with respect to the target report. Within \method, this leads to a mitigation of false-positive increase.

\subsection{Semantic Poisoning}
\label{subsec:adv_semantic_poisoning}

Semantic poisoning targets a different failure mode: \tpobj{} suppresses
recognizable surface cues for genuinely supported entities without introducing
new unsupported behaviors. Recall is therefore the primary diagnostic.
Under this perturbation, \method's recall decreases from $0.635$ to $0.545$,
a $9.0$ percentage-point ($14.2\%$ relative) reduction. Its precision, by
contrast, remains comparatively stable at $0.696$, only $3.6\%$ below its
clean value. The asymmetric change in precision and recall is consistent with
the intended attack objective, as semantic rewriting primarily causes supported
techniques to be missed rather than inducing additional false positives.

\method does not exhibit the smallest \emph{relative} recall degradation under semantic poisoning. Nevertheless, because \method begins from a substantially stronger clean operating point, it still achieves the highest absolute recall on the poisoned reports ($0.545$), compared with $0.511$ for GPT-5.5. Semantic poisoning therefore exposes the principal residual weakness in \method's adversarial robustness: evidence-grounded verification is effective at rejecting unsupported candidates, but cannot fully recover a true technique when semantic obfuscation prevents that candidate or its supporting evidence from
being recognized in the first place.

\subsection{Overall Robustness}
\label{subsec:adv_summary_text}

Across the three adversarial variants, \method incurs a mean relative precision
degradation of $-7.0\%$, the smallest among all evaluated systems and roughly
half that of GPT-5.5 ($-14.0\%$) and GPT-5.4 ($-12.2\%$).
Moreover, \method retains the highest absolute precision under both
\fpobj{} perturbations as well as under semantic poisoning. These results
support \ctag{C3} that reasoning-specialized multi-expert verification makes
document-level extraction substantially less susceptible to plausible but
unsupported or contaminated CTI content.

The precision--recall decomposition also clarifies the boundary of this
robustness. \method is particularly effective against \fpobj{} attacks, where multi-expert verification can contest and reject unsupported candidates after they are proposed.
Under \tpobj{} semantic poisoning, however, weakened behavioral cues can
prevent supported candidates from being recovered at all, producing a larger
recall sensitivity despite the highest remaining absolute recall. Thus,
\method improves robustness to both classes in absolute extraction quality,
while semantic true-positive suppression remains a distinct challenge from
false-positive filtering.

\begin{figure*}[!htbp]
  \centering
  \includegraphics[
    width=\linewidth,
    keepaspectratio,
    trim=0mm 23mm 0mm 0mm,
    clip
  ]{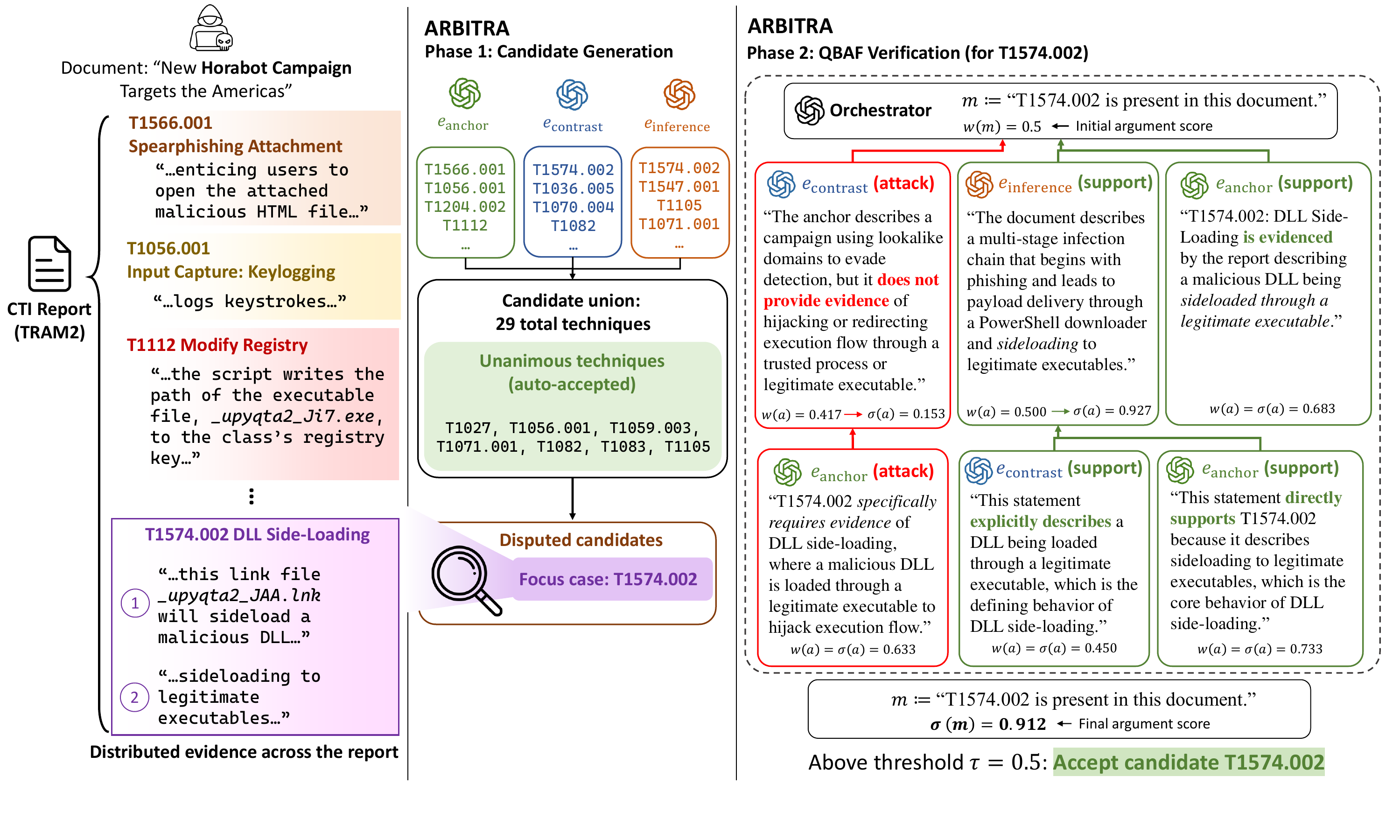}
  \caption{Example case study of \method on the New Horabot campaign CTI report.}
  \label{fig:usecase-fig}
\end{figure*}

We additionally extend the adversarial evaluation using the single-call
verifier model from Section~\ref{subsec:eval_verifier} in
Appendix~\ref{app:adv-verifier} (Table~\ref{tab:abl_adv_verifier}).
The QBAF advantage over single-call verification is preserved across all three perturbation classes under Reasoning~FT in terms of F1.

\section{Case Study}
\label{sec:discussion}

We examine an individual TRAM2 document as a case study with qualitatively distinct extraction challenges and illustrate where \method provides interpretable advantages. A comprehensive workflow of \method is shown in Figure~\ref{fig:usecase-fig}.

The New Horabot campaign report describes a multi-stage phishing and banking trojan operation with overlapping delivery, execution, persistence, credential-access, and exfiltration mechanisms.
\method Reasoning FT achieves $P/R/F1=0.947/0.720/0.818$ on this document, compared with $0.750/0.600/0.667$ for Basic FT and $0.500/0.240/0.324$ for No-FT.
The Reasoning FT candidate union contains 29 techniques and seven unanimous ground-truth techniques, demonstrating that the experts agree on several core behaviors while surfacing additional disputed candidates for verification.

The expert outputs also show why the QBAF layer is useful for distributed evidence.
The evidence-grounding experts locate direct textual anchors such as \textit{``enticing users to open the attached malicious HTML file''} for \texttt{T1566.001} (Phishing: Spearphishing Attachment) and \textit{``logs keystrokes''} for \texttt{T1056.001} (Input Capture: Keylogging).
For \texttt{T1574.002}~\footnote{\texttt{T1574.002} has since been merged into \texttt{T1574.001} in a later ATT\&CK release; we retain \texttt{T1574.002} as the ground-truth label because TRAM2 was annotated against an earlier version of the framework.} (DLL Side-Loading), whose evidence is distributed across paragraphs describing malicious DLLs loaded through legitimate executables, both $e_{\text{anchor}}$ and $e_{\text{inference}}$ independently nominate the technique. Notably, $e_{\text{contrast}}$ produces a false argument claiming that the report contains no evidence for \texttt{T1574.002}. This argument is then directly attacked by $e_{\text{anchor}}$, weakening the original false claim.
The resulting QBAF score remains high (0.912): the discussion contains attacks against weak or mismatched side-loading evidence, but these are outweighed by stronger supports grounded in the actual DLL side-loading chain. Disagreement is thus not simply averaged away but instead converted into explicit support and attack structures, which is the same mechanism that underpins \method's robustness to adversarial content~\ctag{C3} (Section~\ref{sec:evaluation_adversarial}).
\section{Conclusion}
\label{sec:conclusion}

We presented \method, a document-level \cti extraction framework that separates candidate discovery from candidate acceptance through reasoning-specialized multi-expert argumentation. \method first constructs a broad union of expert-proposed ATT\&CK entities and then subjects disputed candidates to evidence-grounded QBAF verification, providing precision control under full-document extraction \ctag{C1}, complementary reasoning for verification \ctag{C2}, and robustness to adversarial report content \ctag{C3}. Across TRAM2 and AnnoCTR, \method achieves the strongest precision-oriented operating point among the evaluated systems, surpasses the sentence-level baseline on all three metrics on TRAM2, and outperforms recent CTI extraction systems under the common technique-level evaluation. Our ablations show that these gains arise from the combination of reasoning-specialized experts and structured multi-expert verification rather than either component alone. Under adversarial perturbations, \method is particularly effective against false-positive injection from conflicting TTP statements and cross-report contamination, limiting mean precision degradation to roughly half that of frontier-model baselines while retaining strong absolute performance.

Several limitations remain. Semantic poisoning exposes greater recall sensitivity when behavioral cues are sufficiently obfuscated to suppress candidates from the union before verification. Per-candidate multi-expert discussion also introduces additional inference latency in exchange for stronger verification. Extending the reasoning corpora with behaviorally equivalent paraphrases and improving recovery of unseen entities are promising directions for future work.

\bibliographystyle{plainurl}
\bibliography{main}

\clearpage
\appendix
\setcounter{subsection}{0}
\renewcommand{\thesubsection}{A.\arabic{subsection}}

\makeatletter
\def\thesubsectiondis{A.\arabic{subsection}}
\makeatother
\section*{Ethical Considerations}

Our work studies automated MITRE \attack extraction from publicly available CTI benchmark datasets and does not involve human subjects or the collection of new private or personally identifiable information. The experiments use the released TRAM2 and AnnoCTR datasets and their existing annotations, and our evaluation is conducted entirely on benchmark CTI documents. Accordingly, the primary ethical considerations arise not from data collection, but from the possible downstream use of automated CTI extraction and from the adversarial perturbations introduced for robustness evaluation.

First, inaccurate \attack mappings can affect downstream security analysis: false positives may direct analysts toward unsupported behaviors, while false negatives may cause relevant techniques to be overlooked. \method is intended as a decision-support mechanism rather than a replacement for analyst judgment. Its per-candidate arguments are designed to make disputed mappings inspectable, but they should not be interpreted as guarantees of correctness. Deployments such as operational CTI pipelines should therefore retain human oversight and validate extracted mappings against the underlying report evidence.

Second, our CTI threat model and the adversarial evaluation consider conflicting TTP statements, cross-report contamination, and semantic rewriting of CTI text. These perturbations could in principle inform attempts to mislead automated intelligence-processing systems. However, the evaluation is limited to text-level manipulation of benchmark reports and does not develop exploits, malware, or techniques for compromising deployed systems. We include these perturbations to characterize realistic failure modes and to support the development of more robust CTI extraction systems. We believe that documenting these weaknesses and evaluating defenses against them provides greater defensive benefit than the limited misuse potential introduced by their disclosure.

\section*{Open Source Availability}

The ARBITRA codebase is not publicly available at the time of the preprint and will be released following acceptance and publication of this work at a peer-reviewed venue.


\subsection{Benchmark Dataset Statistics}
\label{app:data-statistics}
We provide the full dataset statistics for the three benchmark datasets used in the evaluation (TRAM2, AnnoCTR, and CTIBench-ATE) in Table~\ref{tab:benchmark_statistics}. Our statistics show minor discrepancies with those reported in the SoK paper~\cite{sok_ttp_extraction_2025}. These differences stem from sentence tokenization choices and preprocessing details that are not fully specified in the SoK paper. However, our splits are \emph{identical}: we evaluate directly on
the dataset files released as part of their open-science artifact~\cite{sok_ttp_extraction_2025}, so the
statistics reported here reflect recalculations over the exact same documents and annotations. We therefore regard our counts as the more precise reference for the experimental results presented in this paper.

\begin{table}[ht!]
\centering
\caption{%
  Expert pool combination ablation on TRAM2 ($\tau=0.5$). All configurations
  use the same Phase~2 QBAF orchestrator (GPT-4o-mini). The canonical
  \method{} Reasoning~FT pool is repeated from
  Table~\ref{tab:main_results_summary} for reference.
  \textbf{Bold}: best; \textit{italic}: second best per column.%
}
\label{tab:abl_exp_combo}
\small
\setlength{\tabcolsep}{4pt}
\renewcommand{\arraystretch}{1.12}
\resizebox{0.8\columnwidth}{!}{%
\begin{tabular}{@{}l
  >{\centering\arraybackslash}p{0.75cm}
  >{\centering\arraybackslash}p{0.75cm}
  >{\centering\arraybackslash}p{0.75cm}@{}}
\toprule
\textbf{Expert Configuration} & \textbf{P} & \textbf{R} & \textbf{F1} \\
\midrule
\multicolumn{4}{@{}l}{\textit{Canonical (3 diverse reasoning experts)}} \\[1pt]
\quad $e_{\mathrm{anchor}}$, $e_{\mathrm{contrast}}$, $e_{\mathrm{inference}}$
  & 0.722 & \textit{0.635} & \textbf{0.649} \\
\midrule
\multicolumn{4}{@{}l}{\textit{2-expert subsets}} \\[1pt]
\quad $e_{\mathrm{anchor}}$, $e_{\mathrm{inference}}$
  & 0.691 & \textbf{0.644} & 0.640 \\
\quad $e_{\mathrm{anchor}}$, $e_{\mathrm{contrast}}$
  & 0.707 & 0.589 & 0.602 \\
\quad $e_{\mathrm{contrast}}$, $e_{\mathrm{inference}}$
  & \textbf{0.756} & 0.561 & 0.617 \\
\quad $e_{\mathrm{basic}}$, $e_{\mathrm{inference}}$
  & 0.721 & 0.581 & 0.625 \\
\midrule
\multicolumn{4}{@{}l}{\textit{Mixed reasoning + basic FT (3 experts)}} \\[1pt]
\quad $e_{\mathrm{contrast}}$, $e_{\mathrm{inference}}$, $e_{\mathrm{basic}}$
  & 0.692 & 0.635 & \textit{0.643} \\
\midrule
\multicolumn{4}{@{}l}{\textit{Duplicate experts (reduced diversity)}} \\[1pt]
\quad $e_{\mathrm{inference}}$$\,{\times}\,$2, $e_{\mathrm{anchor}}$
  & 0.708 & 0.613 & 0.632 \\
\quad $e_{\mathrm{inference}}$$\,{\times}\,$2, $e_{\mathrm{contrast}}$
  & 0.666 & 0.516 & 0.571 \\
\quad $e_{\mathrm{inference}}$$\,{\times}\,$3
  & \textit{0.736} & 0.563 & 0.624 \\
\quad $e_{\mathrm{anchor}}$$\,{\times}\,$3
  & 0.648 & 0.616 & 0.610 \\
\quad $e_{\mathrm{contrast}}$$\,{\times}\,$2, $e_{\mathrm{anchor}}$
  & 0.727 & 0.487 & 0.562 \\
\quad $e_{\mathrm{contrast}}$$\,{\times}\,$2, $e_{\mathrm{inference}}$
  & 0.726 & 0.503 & 0.572 \\
\bottomrule
\end{tabular}%
}
\end{table}

\begin{table*}[t]
\centering
\caption{%
  Comprehensive corpus and annotation statistics for the three evaluation benchmarks as used in our evaluations.
}
\label{tab:benchmark_statistics}
\renewcommand{\arraystretch}{1.18}
\setlength{\tabcolsep}{5pt}
\resizebox{0.9\textwidth}{!}{%
\begin{tabular}{%
  l                       
  r r r                   
  r r r                   
  r                       
}
\toprule
&
\multicolumn{3}{c}{\textbf{TRAM2}~\cite{tram}} &
\multicolumn{3}{c}{\textbf{AnnoCTR}~\cite{annoctr}} &
\multicolumn{1}{c}{\textbf{CTIBench-ATE}~\cite{ctibench}} \\
\cmidrule(lr){2-4}
\cmidrule(lr){5-7}
\cmidrule(lr){8-8}
\textbf{Statistic} &
\textbf{Train} & \textbf{Test} & \textbf{Full} &
\textbf{Train} & \textbf{Test} & \textbf{Full} &
\textbf{Test} \\

\midrule
\multicolumn{8}{l}{\textit{(a)~Corpus Statistics}} \\[1pt]
 
\quad\# Documents
  & 120     & 31     & 151
  & 83      & 34     & 117
  & 60 \\
 
\quad\# Sentences
  & 15{,}358 & 3{,}820 & 19{,}178
  & 3{,}133  & 2{,}017 & 5{,}150
  & 401 \\
 
\quad Sent.\,/\,Doc\ \ Mean\,$\pm$\,SD
  & $127.98 \pm 88.30$ & $123.23 \pm 61.66$ & $127.01 \pm 83.36$
  & $37.75 \pm 54.63$  & $59.32 \pm 39.29$  & $44.02 \pm 54.74$
  & $6.68 \pm 2.69$ \\
 
\quad Sent.\,/\,Doc\ \ Median
  & 105.5 & 110.0 & 107.0
  & 22.0  & 42.5  & 33.0
  & 6.5 \\
 
\quad Sent.\,/\,Doc\ \ Min\,/\,Max
  & 18\,/\,522 & 2\,/\,256 & 2\,/\,522
  & 1\,/\,463  & 23\,/\,177 & 1\,/\,522
  & 1\,/\,14 \\

\midrule
\multicolumn{8}{l}{\textit{(b)~Unique Ground-Truth Labels by MITRE ATT\&CK Entity Type}} \\[1pt]

\quad Techniques (parent, T\textit{xxxx})
  & 26  & 25  & 26
  & 118 & 97  & 133
  & 115 \\

\quad Sub-Techniques (T\textit{xxxx.yyy})
  & 24  & 23  & 24
  & \multicolumn{3}{c}{---}
  & --- \\

\quad Total (Sub-)Techniques
  & \textbf{50} & \textbf{48} & \textbf{50}
  & \multicolumn{3}{c}{---}
  & --- \\

\quad Tactics (TA\textit{xxxx})
  & \multicolumn{3}{c}{---}
  & 14  & 12  & 14
  & --- \\

\quad Software / Malware (S\textit{xxxx})
  & \multicolumn{3}{c}{---}
  & 57  & 23  & 63
  & --- \\

\quad Threat Groups (G\textit{xxxx})
  & \multicolumn{3}{c}{---}
  & 31  & 13  & 38
  & --- \\

\midrule
\multicolumn{8}{l}{\textit{(c)~Dataset Properties}} \\[1pt]

\quad Annotation Granularity
  & \multicolumn{3}{c}{Sentence-level}
  & \multicolumn{3}{c}{Word-level (span)}
  & Document-level \\

\quad Training Split Available
  & \multicolumn{3}{c}{\checkmark}
  & \multicolumn{3}{c}{\checkmark}
  & --- \\

\quad ATT\&CK Coverage
  & \multicolumn{3}{c}{Top-50 (sub-)techniques}
  & \multicolumn{3}{c}{133 techniques (full set)}
  & 115 techniques (full set) \\

\bottomrule
\end{tabular}%
}
\end{table*}
\subsection{Effect of Expert Pool Composition}
\label{app:abl-exp-combo}

The main evaluation uses a fixed pool of three reasoning-specialized experts
($e_{\mathrm{anchor}}$, $e_{\mathrm{contrast}}$, $e_{\mathrm{inference}}$) chosen to
cover complementary evidential failure modes, as described in
Section~\ref{subsec:method_reasoning_experts}.
In this section, we verify whether this specific composition is necessary, or whether
alternative pools such as dropping one expert, mixing reasoning-FT with basic-FT experts, or using duplicate instances of a single strong expert can achieve comparable performance.
Table~\ref{tab:abl_exp_combo} reports eleven alternative configurations against the canonical pool on TRAM2 at $\tau=0.5$.

The canonical three-expert pool achieves the highest F1 (0.649) across all eleven
alternatives, even though certain configurations exceed it on individual metrics. These gains, however, come at the cost of the complementary metric---high precision
pools suppress recall, and high recall pools admit more false positives---whereas
the three-expert canonical pool achieves the best precision-recall balance as
measured by F1. Among 2-expert subsets, the spread in F1 is substantial, illustrating
that the three experts are not interchangeable: removing any one of them unbalances the evidential coverage that Phase~2 relies on.
The mixed-FT pool ($e_{\mathrm{contrast}}$, $e_{\mathrm{inference}}$, $e_{\mathrm{basic}}$)
is the closest overall alternative, and contains
the two highest-performing standalone experts on TRAM2
(Table~\ref{tab:main_results_summary}; $e_{\mathrm{basic}}$ F1~$=$~0.577,
$e_{\mathrm{inference}}$ F1~$=$~0.546). However, substituting $e_{\mathrm{anchor}}$ with $e_{\mathrm{basic}}$ still yields a weaker pool, showing that Phase~2 benefit derives from reasoning diversity rather than individual model strength.

The duplicate-expert configurations reveal that the severity of recall collapse under
reduced diversity is expert-dependent.
Duplicating $e_{\mathrm{contrast}}$ causes the most severe degradation. Across all duplicate settings, the mechanism is consistent: using the same models causes the QBAF discussion to lose its deliberative character and over-reject or
over-accept candidates depending on that expert's inherent tendency.
\subsection{Adversarial Evaluation Using a Verifier}
\label{app:adv-verifier}

Section~\ref{subsec:eval_verifier} and Table~\ref{tab:ablation_phase2} establish that
replacing \method's Phase~2 QBAF discussion with a single verifier LLM call degrades
performance on clean TRAM2, with the gap growing substantially under Reasoning~FT settings. We follow this up by verifying whether this advantage holds under adversarial conditions:
a single-call verifier sees the same perturbed document as the QBAF discussion and may be differentially susceptible to injected noise or suppressed phrasing.
Table~\ref{tab:abl_adv_verifier} reports this comparison across the three adversarial
variants defined in Section~\ref{subsec:method_adversarial}.

The pattern from the clean evaluation transfers directly to the adversarial setting.
Under \textit{Condition~1} (No fine-tuning), QBAF and the verifier remain effectively
indistinguishable across all three perturbation types, consistent with the clean finding that neither mechanism benefits significantly when Phase~1 experts lack reasoning supervision. Under \textit{Conditions~2 and~3} (Reasoning~FT), QBAF maintains a consistent F1 advantage across every adversarial variant. Notably, the advantage is driven almost entirely by recall, which is also consistent with the behavior observed on clean data in Table~\ref{tab:ablation_phase2}, where single-call verifiers suppress true positives rather than over-accepting false ones. Under adversarial perturbation, this conservative bias is amplified: the verifier encounters a noisier document and grows more skeptical of all candidates.

\begin{table*}[t]
\centering
\caption{%
  Adversarial robustness ablation on three adversarial TRAM2 variants:
  QBAF Discussion (\method{}) vs.\ a single Verifier LLM as the Phase~2
  mechanism. Conditions are detailed in the section headers;
  \method{} (QBAF) values in conditions~2 and~3 are identical and repeated
  for direct comparison. Clean baseline performance is reported in
  Table~\ref{tab:ablation_phase2}.
  $\Delta$ rows show \method{} minus Verifier LLM in percentage points~(pp);
  \textcolor{green!60!black}{green}/\textcolor{red!70!black}{red}~=~QBAF
  advantage/disadvantage. \textbf{Bold}: higher value within each condition block.%
}
\label{tab:abl_adv_verifier}
\scriptsize
\setlength{\tabcolsep}{2.5pt}
\renewcommand{\arraystretch}{1.12}
\resizebox{0.9\linewidth}{!}{%
\begin{tabular}{@{}p{3.0cm} p{2.90cm}
  *{3}{>{\centering\arraybackslash}p{0.82cm}}
  *{3}{>{\centering\arraybackslash}p{0.82cm}}
  *{3}{>{\centering\arraybackslash}p{0.82cm}}@{}}
\toprule
\multirow{2}{*}{\textbf{System}}
  & \multirow{2}{*}{\textbf{Verification Mechanism}}
  & \multicolumn{3}{c}{\textbf{Conflicting}}
  & \multicolumn{3}{c}{\textbf{Cross-report}}
  & \multicolumn{3}{c}{\textbf{Semantic}} \\
\cmidrule(lr){3-5}\cmidrule(lr){6-8}\cmidrule(l){9-11}
  &
  & \textbf{P} & \textbf{R} & \textbf{F1}
  & \textbf{P} & \textbf{R} & \textbf{F1}
  & \textbf{P} & \textbf{R} & \textbf{F1} \\
\midrule

\multicolumn{11}{@{}l}{%
  \textit{Condition 1: No fine-tuning\ \ (Phase~1: 3$\times$ GPT-4o-mini)}%
} \\[1pt]
\method{} (QBAF)
  & QBAF Discussion
  & \textbf{0.286} & 0.295 & 0.273
  & \textbf{0.310} & 0.303 & 0.288
  & \textbf{0.309} & \textbf{0.316} & \textbf{0.286} \\
\method{} (Verifier LLM)
  & Verifier LLM (GPT-4o-mini)
  & 0.274 & \textbf{0.315} & \textbf{0.274}
  & 0.298 & \textbf{0.337} & \textbf{0.297}
  & 0.279 & 0.309 & 0.269 \\[-3pt]
  & {\scriptsize\textit{$\Delta$ QBAF $-$ Verifier}}
  & {\scriptsize\textcolor{green!60!black}{+1.2\,pp}}
    & {\scriptsize\textcolor{red!70!black}{$-$2.0\,pp}}
    & {\scriptsize\textcolor{red!70!black}{$-$0.1\,pp}}
  & {\scriptsize\textcolor{green!60!black}{+1.2\,pp}}
    & {\scriptsize\textcolor{red!70!black}{$-$3.4\,pp}}
    & {\scriptsize\textcolor{red!70!black}{$-$0.9\,pp}}
  & {\scriptsize\textcolor{green!60!black}{+3.0\,pp}}
    & {\scriptsize\textcolor{green!60!black}{+0.7\,pp}}
    & {\scriptsize\textcolor{green!60!black}{+1.7\,pp}} \\[2pt]

\midrule

\multicolumn{11}{@{}l}{%
  \textit{Condition 2: Reasoning FT\ \ (Phase~1: $e_{\mathrm{anchor}}$,
  $e_{\mathrm{contrast}}$, $e_{\mathrm{inference}}$;\ verifier: GPT-4o-mini)}%
} \\[1pt]
\method{} (QBAF)
  & QBAF Discussion
  & \textbf{0.662} & \textbf{0.576} & \textbf{0.595}
  & \textbf{0.657} & \textbf{0.551} & \textbf{0.583}
  & 0.696 & \textbf{0.545} & \textbf{0.576} \\
\method{} (Verifier LLM)
  & Verifier LLM (GPT-4o-mini)
  & 0.644 & 0.475 & 0.528
  & 0.644 & 0.473 & 0.511
  & \textbf{0.720} & 0.440 & 0.530 \\[-3pt]
  & {\scriptsize\textit{$\Delta$ QBAF $-$ Verifier}}
  & {\scriptsize\textcolor{green!60!black}{+1.8\,pp}}
    & {\scriptsize\textcolor{green!60!black}{+10.1\,pp}}
    & {\scriptsize\textcolor{green!60!black}{+6.7\,pp}}
  & {\scriptsize\textcolor{green!60!black}{+1.3\,pp}}
    & {\scriptsize\textcolor{green!60!black}{+7.8\,pp}}
    & {\scriptsize\textcolor{green!60!black}{+7.2\,pp}}
  & {\scriptsize\textcolor{red!70!black}{$-$2.4\,pp}}
    & {\scriptsize\textcolor{green!60!black}{+10.5\,pp}}
    & {\scriptsize\textcolor{green!60!black}{+4.6\,pp}} \\[2pt]

\midrule

\multicolumn{11}{@{}l}{%
  \textit{Condition 3: Reasoning FT\ \ (Phase~1: $e_{\mathrm{anchor}}$,
  $e_{\mathrm{contrast}}$, $e_{\mathrm{inference}}$;\ verifier: $e_{\mathrm{best}}$)}%
} \\[1pt]
\method{} (QBAF)
  & QBAF Discussion
  & 0.662 & \textbf{0.576} & \textbf{0.595}
  & 0.657 & \textbf{0.551} & \textbf{0.583}
  & \textbf{0.696} & \textbf{0.545} & \textbf{0.576} \\
\method{} (Verifier LLM)
  & Verifier LLM ($e_{\mathrm{best}}$)
  & \textbf{0.698} & 0.416 & 0.501
  & \textbf{0.677} & 0.400 & 0.469
  & 0.685 & 0.364 & 0.459 \\[-3pt]
  & {\scriptsize\textit{$\Delta$ QBAF $-$ Verifier}}
  & {\scriptsize\textcolor{red!70!black}{$-$3.6\,pp}}
    & {\scriptsize\textcolor{green!60!black}{+16.0\,pp}}
    & {\scriptsize\textcolor{green!60!black}{+9.4\,pp}}
  & {\scriptsize\textcolor{red!70!black}{$-$2.0\,pp}}
    & {\scriptsize\textcolor{green!60!black}{+15.1\,pp}}
    & {\scriptsize\textcolor{green!60!black}{+11.4\,pp}}
  & {\scriptsize\textcolor{green!60!black}{+1.1\,pp}}
    & {\scriptsize\textcolor{green!60!black}{+18.1\,pp}}
    & {\scriptsize\textcolor{green!60!black}{+11.7\,pp}} \\

\bottomrule
\end{tabular}%
}
\end{table*}
\subsection{Cross-Benchmark Transfer to CTIBench-ATE}
\label{app:cross-transfer}

In this section, we investigate whether fine-tuned experts can transfer across benchmarks: whether experts
trained on TRAM2 or AnnoCTR have learned generalizable TTP extraction skills and can be applied to a held-out dataset with a different label space without any benchmark-specific retraining. We test this directly using CTIBench-ATE~\cite{ctibench} as the transfer target.

Table~\ref{tab:results_ctibench_ate} reports zero-shot transfer results on CTIBench-ATE,
where all \method{} configurations use expert pools trained on either TRAM2 or AnnoCTR
with no CTIBench-ATE-specific fine-tuning.
In this evaluation, No-FT configurations outperform all fine-tuned variants:
\method{} No-FT ($3{\times}$GPT-4.1-mini) achieves the best overall performance. Fine-tuned expert pools, by contrast, consistently degrade relative to their No-FT counterparts. The primary explanation is label space mismatch.
CTIBench-ATE covers 115 distinct ATT\&CK techniques drawn from the full Enterprise matrix,
while TRAM2 fine-tuning targets only 50 (sub-)techniques and AnnoCTR covers a
partially overlapping set of 133 parent techniques
(see Table~\ref{tab:benchmark_statistics}).
Fine-tuned experts develop strong priors toward the techniques represented in their
training data; on CTIBench-ATE documents, this results in recall degradation as experts decline to predict identifiers outside their domain.

Although fine-tuned pools degrade under the label-space shift, the orchestration mechanism still provides measurable benefit in the No-FT setting (+5.0--+7.3 pp F1 over the corresponding standalone baselines), showing that the multi-expert verification pipeline itself transfers across benchmarks without benchmark-specific tuning. What the cross-benchmark results expose instead is a fundamental constraint
of the CTI dataset ecosystem: as Büchel et al.\ conclude in their SoK study, data-driven
TTP extraction \textit{is fundamentally limited by the lack of manually annotated
datasets}~\cite{sok_ttp_extraction_2025}.
Our results also support this statement by showing that fine-tuned experts that achieve strong in-distribution performance regress sharply with target label space shifts.

\begin{table*}[t]
\centering
\caption{Zero-shot transfer results on CTIBench-ATE. Backbone models are indicated in parentheses. \method{} applies no CTIBench-ATE-specific tuning. Percentage-point deltas~(pp) are relative to the corresponding standalone baseline: the single LLM for No-FT rows, $e_{\mathrm{basic}}$ for Basic FT rows, and the best-F1 expert ($e_{\mathrm{anchor}}$ for TRAM2; $e_{\mathrm{contrast}}$ for AnnoCTR) for Reasoning FT rows. \textbf{Bold}: best per column.}
\label{tab:results_ctibench_ate}
\footnotesize
\setlength{\tabcolsep}{4pt}
\renewcommand{\arraystretch}{1.12}
\resizebox{0.85\linewidth}{!}{%
\begin{tabular}{@{}p{4.0cm} p{7.2cm} >{\centering\arraybackslash}p{1.0cm} >{\centering\arraybackslash}p{1.0cm} >{\centering\arraybackslash}p{1.0cm}@{}}
\toprule
\textbf{System} & \textbf{Expert Config} & \textbf{P} & \textbf{R} & \textbf{F1} \\
\midrule
\multicolumn{5}{@{}l}{\textit{Document-level LLM baselines}} \\
GPT-4o-mini
  &
  & 0.446 & 0.398 & 0.416 \\
GPT-4.1-mini
  &
  & 0.603 & 0.609 & 0.597 \\
\midrule
\multicolumn{5}{@{}l}{\textit{Finetuned expert instances — trained on TRAM2, transferred to CTIBench-ATE}} \\
$e_{\mathrm{basic}}$ {\scriptsize (finetuned GPT-4o-mini)}
  &
  & 0.510 & 0.381 & 0.429 \\
$e_{\mathrm{anchor}}$ {\scriptsize (finetuned GPT-4.1-mini)}
  &
  & 0.469 & 0.408 & 0.427 \\
$e_{\mathrm{contrast}}$ {\scriptsize (finetuned GPT-4o-mini)}
  &
  & 0.269 & 0.295 & 0.271 \\
$e_{\mathrm{inference}}$ {\scriptsize (finetuned GPT-4o-mini)}
  &
  & 0.428 & 0.310 & 0.338 \\
\midrule
\multicolumn{5}{@{}l}{\textit{Finetuned expert instances — trained on AnnoCTR, transferred to CTIBench-ATE}} \\
$e_{\mathrm{basic}}$ {\scriptsize (finetuned GPT-4.1-mini)}
  &
  & 0.358 & 0.294 & 0.315 \\
$e_{\mathrm{anchor}}$ {\scriptsize (finetuned GPT-4.1-mini)}
  &
  & 0.413 & 0.272 & 0.322 \\
$e_{\mathrm{contrast}}$ {\scriptsize (finetuned GPT-4.1-mini)}
  &
  & 0.431 & 0.273 & 0.328 \\
$e_{\mathrm{inference}}$ {\scriptsize (finetuned GPT-4.1-mini)}
  &
  & 0.394 & 0.249 & 0.298 \\
\midrule
\multicolumn{5}{@{}l}{%
  \begin{tabular}[t]{@{}l@{}}
  \textit{Multi-expert orchestration (ours)}\\[-0.15em]
  {\scriptsize Expert pool: GPT-4o-mini / GPT-4.1-mini}
  \end{tabular}%
} \\
\method{}
  & No-FT (3$\times$ GPT-4o-mini)
  & 0.565 & 0.447 & 0.489 \\[-3pt]
  & & {\scriptsize\textcolor{green!60!black}{+11.9\,pp}}
    & {\scriptsize\textcolor{green!60!black}{+4.9\,pp}}
    & {\scriptsize\textcolor{green!60!black}{+7.3\,pp}} \\[2pt]
\method{}
  & No-FT (3$\times$ GPT-4.1-mini)
  & \textbf{0.665} & \textbf{0.649} & \textbf{0.647} \\[-3pt]
  & & {\scriptsize\textcolor{green!60!black}{+6.2\,pp}}
    & {\scriptsize\textcolor{green!60!black}{+4.0\,pp}}
    & {\scriptsize\textcolor{green!60!black}{+5.0\,pp}} \\[2pt]
\method{}
  & Basic FT (3$\times$ $e_{\mathrm{basic}}$; TRAM2)
  & 0.527 & 0.351 & 0.410 \\[-3pt]
  & & {\scriptsize\textcolor{green!60!black}{+1.7\,pp}}
    & {\scriptsize\textcolor{red!70!black}{$-$3.0\,pp}}
    & {\scriptsize\textcolor{red!70!black}{$-$1.9\,pp}} \\[2pt]
\method{}
  & Basic FT (3$\times$ $e_{\mathrm{basic}}$; AnnoCTR)
  & 0.335 & 0.326 & 0.321 \\[-3pt]
  & & {\scriptsize\textcolor{red!70!black}{$-$2.3\,pp}}
    & {\scriptsize\textcolor{green!60!black}{+3.2\,pp}}
    & {\scriptsize\textcolor{green!60!black}{+0.6\,pp}} \\[2pt]
\method{}
  & Reasoning FT ($e_{\mathrm{anchor}}$, $e_{\mathrm{contrast}}$, $e_{\mathrm{inference}}$; TRAM2)
  & 0.490 & 0.300 & 0.360 \\[-3pt]
  & & {\scriptsize\textcolor{green!60!black}{+2.1\,pp}}
    & {\scriptsize\textcolor{red!70!black}{$-$10.8\,pp}}
    & {\scriptsize\textcolor{red!70!black}{$-$6.7\,pp}} \\[2pt]
\method{}
  & Reasoning FT ($e_{\mathrm{anchor}}$, $e_{\mathrm{contrast}}$, $e_{\mathrm{inference}}$; AnnoCTR)
  & 0.520 & 0.256 & 0.331 \\[-3pt]
  & & {\scriptsize\textcolor{green!60!black}{+8.9\,pp}}
    & {\scriptsize\textcolor{red!70!black}{$-$1.7\,pp}}
    & {\scriptsize\textcolor{green!60!black}{+0.3\,pp}} \\
\bottomrule
\end{tabular}%
}
\end{table*}

\end{document}